\documentclass{article} 
\usepackage{preprint,newtxtext}
\usepackage{amsthm}
\newtheorem{theorem}{Theorem}[section]
\usepackage{array}

\usepackage{amsmath,amsfonts,bm}

\def\eqref#1{equation~\ref{#1}}

\def\1{\bm{1}}

\DeclareMathAlphabet{\mathsfit}{\encodingdefault}{\sfdefault}{m}{sl}
\SetMathAlphabet{\mathsfit}{bold}{\encodingdefault}{\sfdefault}{bx}{n}

\newcommand{\E}{\mathbb{E}}

\usepackage{booktabs,longtable}

\usepackage{hyperref}
\hypersetup{colorlinks=true,linkcolor=GSBRIndigo,
  citecolor=GSBRIndigo,urlcolor=GSBRBronze,
  pdftitle={GSBR: Geometric Support Behavior Regularization for Online RL},
  pdfauthor={Charles Xu, Jacob Berg, Liyiming Ke, Sergey Levine}}
\usepackage{url}
\usepackage{algorithm,algpseudocode}
\usepackage{graphicx}
\usepackage{xcolor}
\usepackage{wrapfig}
\usepackage{placeins}
\usepackage{needspace}
\usepackage{booktabs}
\title{GSBR: Geometric Support\\Behavior Regularization for Online RL}

\author{%
  \textbf{Charles Xu\textsuperscript{1}, Jacob Berg\textsuperscript{1},
  Liyiming Ke\textsuperscript{2}, Sergey Levine\textsuperscript{1,2}}\\[5pt]
  {\fontsize{10}{12}\selectfont
  \textsuperscript{1}\,UC Berkeley\qquad
  \textsuperscript{2}\,Physical Intelligence}
}
\date{}

\DeclareRobustCommand{\ours}{\textcolor{GSBRBronze}{\textbf{GSBR}}}

\usepackage{color}

\begin{document}

\maketitle

\begin{abstract}
Regularizing online reinforcement learning (RL) toward a pretrained base policy can improve sample efficiency and learning stability, but regularization that is too restrictive limits how much the policy can improve. Base policies -- typically diffusion or flow-matching models trained by imitation learning on large, diverse datasets -- capture many behaviors that are reasonable but suboptimal. Existing regularizers using such base policies are too restrictive in one of two ways: behavioral cloning regularizers pull the actor toward the full base-policy distribution, penalizing it for discarding suboptimal modes, while hard support constraints forbid improvements that lie just outside the base policy's support. We propose Geometric Support Behavior Regularization (\ours{}), which penalizes actions by their distance from the base policy’s support. Actions within the support incur no penalty, so the actor can concentrate on high-value modes; outside the support, where value estimates are less reliable, the penalty grows smoothly with distance while still permitting improvement when the value gained outweighs the penalty. \ours{} estimates this distance from a finite set of base-policy samples, so it requires only sample access to the base policy. 
Across 19 tasks on four robotic benchmarks, \ours{} raises the average
success rate from 33\% for the base policy to 95\%, achieving the highest
average success rate and throughput among all compared methods, with
about 23\% higher normalized throughput than the next-best RL method.
Project website and open-source code: \url{https://charlesxu0124.github.io/gsbr/}.

\end{abstract}

\begin{figure}[H]
    \centering
    \vspace{-1em}
    \includegraphics[width=1\linewidth]{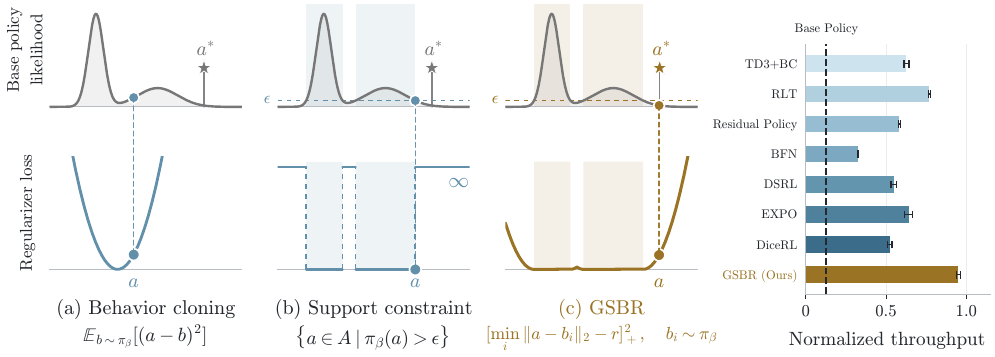}
    \caption{\textbf{Left: }(a) Behavior cloning regularizer matches the base policy distribution, not allowing the RL actor to drop poor modes. (b) Support constraint allows the actor to select the best behavior in the base policy distribution, but excludes improvements beyond it. (c) \textbf{\ours} applies distance-based regularization on deviation outside the base policy support, allowing the actor to discard suboptimal behaviors and guide exploration beyond the prior. \textbf{Right:} \ours{} achieves 23\% higher normalized throughput across 19 tasks and 4 benchmarks than previous methods. 
    }
    \label{fig:teaser}
    
\end{figure}

\section{Introduction}

Online reinforcement learning (RL) can be costly when every interaction happens on a physical system, so practical methods often rely on a behavior prior that focuses exploration on plausible actions. In robotics, such priors are now available at scale: vision-language-action (VLA) and world-action models (WAMs) learn diverse behaviors from large demonstration datasets~\citep{kim2024openvla,black2024pi0,ye2026dreamzero}, but often lack the robustness and throughput required for reliable deployment~\citep{physicalintelligence2025recap,xu2026rlt}. 
RL fine-tuning can close this gap by updating a policy that is regularized toward the pretrained base model.
This raises a general question for RL: \textit{how should a behavior regularizer preserve the base policy's guidance while giving the actor enough freedom to improve on it?}

Improving upon base policies typically has two requirements: mode selection and local refinement. Consider a scene with two objects and the task is to grasp a specified target. Foundation models trained on large-scale data often produce diverse behaviors - and such base policies might have two action modes, one to grasp each object. The actor should be updated to discard the mode that reaches for the wrong object; we call this \emph{mode selection}.
However, existing behavior constraints can hinder one of the requirements and restrict policy improvement. 
Action regression~\citep{fujimoto2021minimalist,park2025fql} and distribution-matching regularizers~\citep{li2026qam,fang2025dac} tie the actor to the base policy's full action distribution, penalizing the actor even for discarding poor modes (Figure~\ref{fig:teaser}(a)).
Support constraints~\citep{kumar2019bear,wagenmaker2025dsrl,li2025qc} remove this penalty and permit mode selection, but can be too restrictive and often exclude optimal actions outside the base policy's support (Figure~\ref{fig:teaser}(b)). 

Our key insight is to allow both mode selection and local refinement by introducing a geometric relaxation of the support constraint.
We introduce Geometric Support Behavior Regularization (\ours), an RL algorithm that estimates the support of the base policy using finite samples, and regularizes actions differently based on whether they are inside or outside the support.
Actions inside the estimated support incur no penalty, allowing the actor to discard poor modes and concentrate on better ones without preserving their relative frequencies. Outside the support, \ours{} replaces the infinite penalty of a hard constraint with a finite one that grows with distance, reflecting that value estimates become less reliable far from the base policy. Because the support of a learned base policy is intractable to compute, we approximate it by the union of neighborhoods around actions sampled from the base policy, with a radius that adapts to the observed samples' distances from each other. \ours{} provides a practical sample-based geometric support regularizer fitting for the diverse distribution of large-scale robot foundation policies.

Our contributions are (1) a geometric behavior regularizer that permits both mode selection and local refinement in RL finetuning, (2) a sample-based approximation of multi-modal action distribution with an adaptive radius, with error bounds and a guarantee that the error vanishes as the number of reference samples grows, (3) an evaluation across 19 tasks drawn from four benchmarks where \ours{} raises the average
success rate from 33\% for the base policy to 95\%, achieving the highest
average success rate and throughput among all compared methods, with
about 23\% higher normalized throughput than the next-best RL method.
Further analysis shows that our approximation closely matches the true support region and examines the effects of sample count and support radius.

\section{Related Work}

\label{sec:related-work}

\paragraph{Sample-efficient online RL.}
Learning from costly environment interaction is a central challenge in robotic RL.
Off-policy actor--critic methods address this through experience reuse,
high update-to-data ratios, and critic regularization to improve
data efficiency~\citep{haarnoja2018sac,chen2021redq,hiraoka2022droq}.
Prior data and demonstrations can also help by initializing policies and values,
supplementing replay buffer, and guiding exploration~\citep{ball2023rlpd,nakamoto2023calql,hu2023ibrl}.
These algorithmic advances enable practical systems to learn contact-rich manipulation 
within hours of robot interaction~\citep{luo2024serl,luo2025hilserl,dong2026expoft}.
 \ours{} \textbf{is complementary to these advances}: it further improves sample efficiency through a behavior regularizer that uses a frozen base policy as a prior, reducing unnecessary exploration while allowing policy improvement beyond the prior's support.

\paragraph{RL with generative models, diffusion and flow policies.}
Diffusion and flow policies, including VLAs, represent multimodal
behavior, but they require iterative action generation and complicate RL optimization.
One family updates the generative model itself, through policy
gradients, critic gradients, or denoising
supervision~\citep{ren2024dppo,wang2023diffusionql,li2026qam,lu2025vlarl},
or through distillation and advantage
conditioning~\citep{park2025fql,chen2025conrft,mark2024parl,physicalintelligence2025recap}.
Another keeps the model frozen and trains a lightweight actor that corrects its actions~\citep{silver2018residual,johannink2019residual,ankile2024resip,xiao2025pld,xu2026rlt},
steers its latent noise~\citep{wagenmaker2025dsrl}, or selects among sampled candidates~\citep{dong2026expo,dong2026expoft,sun2026dicerl}, sometimes distilling the result back into the base model~\citep{xiao2025pld,pi2026pi07}.
\ours{} is most similar to the second family; methods within it differ mainly in how the actor is tied to the base policy, which we compare next.

\paragraph{Behavior constraints for RL.}
Behavior constraints guide policy improvement using available experience or a pretrained policy, but differ in how they restrict mode selection and departures from prior behavior.

\par\smallskip
\noindent\textit{Distribution matching and action regression.} 
KL regularization penalizes reallocating probability toward higher-value modes~\citep{wu2019brac,fang2025dac,li2026qam}, and
regression toward sampled actions favors their
mean~\citep{fujimoto2021minimalist,tarasov2023rebrac}.
FQL and RLT~\citep{park2025fql,xu2026rlt} alleviate this mode collapsing problem by explicitly conditioning on noise or reference-action, but their pairwise regularization still ties the actor to the
modes of randomly drawn inputs.
In contrast, \ours{} allows the actor to focus on promising modes and drop poor behaviors (mode selection) without retaining a particular action pairing.

\par\smallskip
\noindent\textit{Support constraints and approximations.} 
Ideal support constraints allow the actor to choose freely among actions with non-negligible density under the behavior prior. They permit mode selection but block local refinement.
BEAR, SPOT, and PRDC~\citep{kumar2019bear,wu2022spot,ran2023prdc} constrain the actor to dataset actions via sample-based MMD or a
density estimator. However, they require access to the dataset and do not leverage a pretrained foundation model's broader behavioral repertoire.
DSRL~\citep{wagenmaker2025dsrl} acts in the base
policy's latent-noise space, while best-of-$N$ methods select the highest-value action among sampled actions~\citep{nakamoto2024vgps,li2025qc,hansenestruch2023idql}.
Both enable mode selection, but remain suboptimal when local refinement is needed.
These limitations motivate \ours{}, which constructs an unpenalized support region from the base policy distribution and applies a finite geometric penalty beyond it, allowing online exploration to depart from the prior's support.

\par\smallskip
\noindent\textit{Residual reinforcement learning.} 
Residual policies allow local refinement, but are often bounded by a prescribed scale~\citep{silver2018residual,johannink2019residual,ankile2024resip}.
BCQ,  EXPO and EXPO-FT~\citep{fujimoto2019bcq,dong2026expo, dong2026expoft} combine bounded corrections with value-based candidate selection to have both mode selection and local refinement.
However, they only allow local refinement up to a \emph{fixed
magnitude} set as a hyperparameter, bounding the maximum allowed deviation without any guidance from the base policy. They can be interpreted as an implicit support constraint that admits an enlarged support region by a \emph{fixed radius}.
DICE-RL replaces the fixed magnitude by using a selectively dropped out squared-distance regularizer on the edit policy~\citep{sun2026dicerl}. While this method can be viewed as an implicit support constraint with a geometric penalty for deviation, it creates constant optimization pressure that pulls even small edits towards each base policy sample.
In contrast, \ours{} explicitly approximates the support region for the base policy distribution, permitting in-distribution action to have zero behavior penalty, enabling free mode selection and refinement within the support region while allowing departures at a finite, distance-dependent cost.

\section{Preliminaries}
\label{sec:preliminaries}

\paragraph{Reinforcement learning.}
We consider a discounted Markov Decision Process (MDP)
$\mathcal{M}=(\mathcal{S},\allowbreak\mathcal{A},\allowbreak P,\allowbreak r,\allowbreak\rho,\allowbreak\gamma)$ with states
$s\in\mathcal{S}$, actions $a\in\mathcal{A}\subseteq\mathbb{R}^d$,
transition distribution $P(\cdot\mid s,a)$, reward $r(s,a)$,
initial-state distribution $\rho$, and discount factor $\gamma\in[0,1)$.
The goal is to learn a policy $\pi_\theta$ that maximizes the expected
discounted return
\begin{equation}
\textstyle
J(\theta)=\mathbb{E}_{\rho,P,\pi_\theta}
[\sum_{t=0}^{\infty}\gamma^t r(s_t,a_t)].
\label{eq:rl-objective}
\end{equation}

\paragraph{Online RL with a pretrained behavior prior.} In designing RL algorithms to interact with the MDP environment and optimize for the return, we leverage a behavior prior to guide exploration and training. 
We study online policy improvement with a frozen pretrained
\emph{base policy} $\pi_\beta$ and an offline transition dataset
$\mathcal{D}_{\mathrm{off}}$ of \emph{task-specific transitions}, e.g., demonstrations or base-policy rollouts. The learner collects additional experience through environment
interaction and trains on both offline and online transitions.  The base policy stays frozen and is accessed only through samples $b\sim\pi_\beta(\cdot\mid s)$; we require no access to its parameters or gradients, so $\pi_\beta$ can be any generative model.

\paragraph{Challenges.} Three properties make this setting difficult:
(1) $\pi_\beta$ is typically multimodal and its modes can differ widely in value, which behaviors are worth keeping depends on the task.
(2) the base action support is intractable: we can draw samples but cannot enumerate all plausible actions (e.g., too expensive to enumerate for every state).
(3) the critic is trained on limited data, its estimates are less reliable when the input is out-of-distribution; an actor that maximizes $Q_\phi$ without restraint will exploit these errors.

\paragraph{Support and optimality assumption about the base policy.}
Let $\pi_\beta(a\mid s)$ denote the base policy's conditional action density.
We define the support of $\pi_\beta$ at
state $s$:
\begin{equation}
S_\beta(s)=\overline{\bigl\{a\in\mathcal{A}:\pi_\beta(a\mid s) > 0\bigr\}},
\label{eq:support}
\end{equation}
and call its elements \emph{in-distribution actions}.
We assume only that the in-distribution actions include reasonable behaviors
for the task, not that they include the optimal action.
We further assume that improvements over $\pi_\beta$ can be obtained in
two ways: by concentrating on the most valuable supported actions
regardless of how often $\pi_\beta$ samples them, which we call
\emph{mode selection}, and by refining actions near the support, which
we call \emph{local refinement}.

\paragraph{Behavior-regularized actor--critic RL.} Following classic actor-critic RL algorithms, we will learn two functions: actor $\pi_\theta(\cdot\mid s,\mathcal{B})$ and critic $Q^{\pi_\theta}(s,a)$. 
The actor $\pi_\theta(\cdot\mid s,\mathcal{B})$ conditions on the state and on a set of $N$ reference actions $\mathcal{B}=\{b_1,\ldots,b_N\}\sim\pi_\beta(\cdot\mid s)^N$.
The critic function $Q^{\pi_\theta}(s,a)$ is the expected discounted return after taking $a$ in $s$ and subsequently following
$\pi_\theta$. We learn it with classic
temporal-difference learning loss,
\begin{equation}
\textstyle
\mathcal{L}_{\mathrm{TD}}(\phi)
= \mathbb{E}_{\substack{(s,a,r,s')\sim\mathcal{D},\;
\mathcal{B}'\sim\pi_\beta(\cdot\mid s')^N,
a'\sim\pi_\theta(\cdot\mid s',\mathcal{B}')}}
\Bigl[\bigl(Q_\phi(s,a)-\operatorname{sg}\bigl(r+\gamma \,
Q_{\bar\phi}(s',a')\bigr)\bigr)^2\Bigr],
\label{eq:td-loss}
\end{equation}
where $\mathcal{D}$ is the replay distribution over offline and online
transitions,
$\operatorname{sg}$ denotes stop-gradient, $Q_{\bar\phi}$ is a slowly updated target critic.
The actor is trained with a behavior-regularized Q-maximization
objective,
\begin{equation}
\mathcal{L}_{\mathrm{actor}}(\theta)
= \mathbb{E}_{\substack{s\sim\mathcal{D},\;
\mathcal{B}\sim\pi_\beta(\cdot\mid s)^N,
a\sim\pi_\theta(\cdot\mid s,\mathcal{B})}}
\bigl[-Q_\phi(s,a)+\alpha\,\ell_{\mathrm{beh}}(a,\mathcal{B})\bigr],
\label{eq:actor-objective-prelim}
\end{equation}
where $\ell_{\mathrm{beh}}$ is a behavior regularization penalty and
$\alpha\geq0$ controls its strength.
The choice of $\ell_{\mathrm{beh}}$ determines which improvements the
actor can reach. 

\section{Geometric Support Behavior Regularizer (\ours)}
\label{sec:method}

\ours{} uses the base policy to identify where exploration is promising while leaving the actor free to select and improve behaviors in that region.
Section~\ref{sec:geometric-support} formulates the desired behavior constraint ($\ell_{\mathrm{beh}}$ in
Equation~\ref{eq:actor-objective-prelim}) as a geometric relaxation of support-constrained policy improvement.
Section~\ref{sec:tube-construction} approximates this penalty from
base-policy samples and bounds the approximation error.
Algorithm~\ref{alg:tube} describes the practical instantiation of our algorithm.

\subsection{Geometric relaxation of support constraints}
\label{sec:geometric-support}
The assumption in Section~\ref{sec:preliminaries} translates into two
requirements on $\ell_{\mathrm{beh}}$.
For mode selection, the penalty should be zero for in-distribution
actions, so that the actor can concentrate on the most valuable modes without preserving the base policy's relative mode frequencies.
For local refinement, actions outside the support should incur a finite penalty rather than being excluded, and this penalty should grow with distance: the assumption places improvements near the support, and the
critic is least reliable far from it.
Together, these requirements call for a region of zero penalty
surrounded by a finite, distance-dependent penalty (Figure~\ref{fig:method}).

Following the definition of support region in Equation~\ref{eq:support}, the true geometric relaxation of the support constraint is given by
\begin{equation}
    \ell_{\mathrm{supp}}(s,a)
    = d(a,S_\beta)^2,
    \qquad
    d(a,\mathcal{X})
    = \inf_{x\in\mathcal{X}}\lVert a-x\rVert_2.
    \label{eq:ideal-support-penalty}
\end{equation}
The penalty is zero throughout $S_\beta$, leaving value maximization free to determine which supported behaviors to favor.
Outside the region, it penalizes the squared distance to the nearest supported action, allowing departures when their predicted value gains justify the cost.
This geometry retains guidance from the prior without imposing a hard boundary on policy improvement.

\begin{methodinsight}
The ideal penalty also admits a Wasserstein interpretation.
Let $\mathcal{P}_2(S_\beta)$
denote the set of distributions with finite second moments supported entirely within $S_\beta$.
For an actor distribution $\pi_\theta(\cdot \mid s)$ with finite second moment,
\begin{equation}
\mathbb{E}_{a \sim \pi_\theta(\cdot \mid s)}
\left[d\!\left(a,S_\beta\right)^2\right]
=
\inf_{\nu \in \mathcal{P}_2(S_\beta)}
W_2^2\!\left(\pi_\theta(\cdot \mid s), \nu\right).
\label{eq}
\end{equation}
The expected penalty is thus the squared Wasserstein distance from the actor distribution to the set of distributions supported in ($S_\beta$), that is, the minimum quadratic transport cost required to move the actor’s probability mass into the support. 
Because $\nu$ can be any distribution on the support, the penalty constrains where the actor places its mass but not how it divides that mass among modes, which is the formal counterpart of mode selection.
\end{methodinsight}

\subsection{Approximating support with reference-action neighborhoods}
\label{sec:tube-construction}
\begin{figure}
    \centering
    
    \includegraphics[width=1\linewidth]{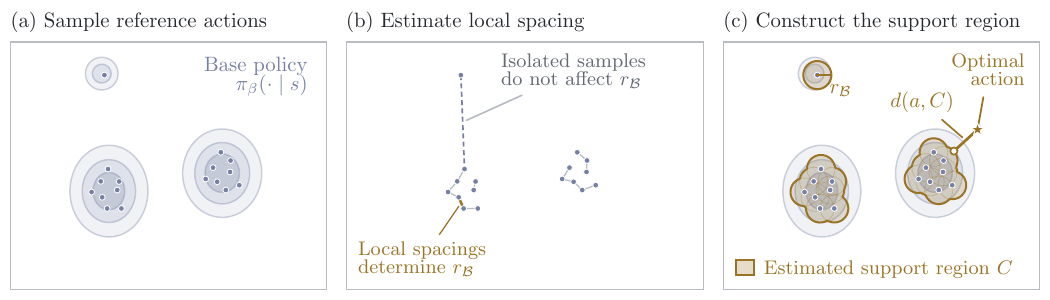}
    
    \caption{\textbf{\ours}. To implement the geometric support regularizer on a generative robot policy, draw samples from $\pi_\beta(\cdot\mid s)$ and compute the nearest-neighbor distance between them. Set $r_\mathcal{B}$ to a quantile over these distances and construct the support region $C$ as the union of balls of radius $r_\mathcal{B}$ around each sample. The behavior regularizer loss is computed as the Euclidean distance to the closest point in $C$.}
    
    \label{fig:method}
\end{figure}

Computing $\ell_{\mathrm{supp}}$ requires the support $S_\beta(s)$,
which is unavailable when the base policy provides only samples. We therefore estimate the support from the reference set $\mathcal{B}$.
However, the reference set alone is a poor estimate: it leaves gaps between plausible actions, and penalizing distance directly to this set would pull the actor toward individual samples.
We address this sampling sparsity by introducing a support ball around each reference action, with an adaptive radius determined by the observed spaces among sampled actions (Figure~\ref{fig:method}).

Given $\mathcal{B}=\{b_1,\ldots,b_N\}$, we compute each reference action's nearest-neighbor distance, and define the support radius as:
\[
u_i:=\min_{j\ne i}\|b_i-b_j\|_2,
\qquad
r_\mathcal{B}:=\operatorname{Quantile}_q
\left(\{u_i\}_{i=1}^{N}\right)
\]

The distance $u_i$ measures the local spacing of reference actions within the same mode or between adjacent action modes; the quantile $q\in(0, 1)$ is a hyperparameter that filters out large distances of isolated action samples. 
The estimated support is the union of closed balls of radius $r_\mathcal{B}$ centered
around each reference action:
\begin{equation}
\begin{aligned}
    C
    &:=
    \bigcup_{i=1}^{N}
    \overline{B}\!\left(
        b_i,r_\mathcal{B}
    \right)
    =
    \left\{
        a\in\mathcal{A} :
        \min_i \lVert a-b_i\rVert_2
        \leq r_\mathcal{B}
    \right\}.
\end{aligned}
\label{eq:empirical-tube}
\end{equation}
Given this estimated support region $C$, we approximate the geometric relaxation of the support constraint as
\begin{equation}
    \ell_{\mathrm{GSBR}}(a,\mathcal{B})
    :=  d(a,C)^2 = \left[
        \min_i ||a-b_i||_2
        -r_\mathcal{B}
      \right]_+^2
    \label{eq:tube-loss}
\end{equation}
where $[x]_+=\max(x,0)$. We train the actor with Equation~\ref{eq:actor-objective-prelim}  by setting $\ell_{\mathrm{beh}} = \ell_{\mathrm{GSBR}}$.

Because $\ell_{\mathrm{GSBR}}$ depends on $\mathcal{B}$ only through the nearest reference, an actor that observes all of $\mathcal{B}$ can move to any sampled mode at zero cost.
This distinguishes \ours{} from regularizers that tie each action to a single sampled reference action~\citep{park2025fql,xu2026rlt}, which retain the
base policy's relative mode frequencies.

\paragraph{Approximation Guarantee.}
The following theorem shows that $\ell_\mathrm{GSBR} +2Dr$ upper bounds the true loss $\ell_{supp}$; their expected difference is bounded and approaches zero as the number of reference samples grows. 

\begin{theorem}[Expected approximation error of the \ours{} loss]
\label{thm:tube-error}
Assume the action space $A$ is compact with diameter \(D\), and that $\pi_\beta(a \mid s)$ has continuous density.
Fix any $a \in A$, and define our reference action set $\mathcal{B}$ as above.

Then, for any radius $r>0$, define a $\lambda_r > 0$ as:
\begin{equation*}
\lambda_r := \inf_{c\in S_\beta} \Pr_{b\sim \pi_\beta}\bigl(\|b-c\|_2\le r\bigr)
\end{equation*}representing the worst-case probability over our support set that a sampled action from $\pi_\beta$ falls within distance $r$ of a given action in $S_\beta$.

The following bounds then hold: 
\begin{align}
\ell_{\mathrm{supp}}(a)
&\le \ell_\mathrm{GSBR}(a,\mathcal B)+2Dr,
\label{eq:tube-one-sided}\\
\mathbb E_\mathcal{B}\!\left[
  \left|\ell_\mathrm{GSBR}(a,\mathcal B)
       -\ell_{\mathrm{supp}}(a)\right|
\right]
&\le 2Dr+D^2(1-\lambda_r)^N.
\label{eq:tube-expected-error}
\end{align}
The upper bound in Equation~\ref{eq:tube-one-sided}
also holds with $r=r_\mathcal{B}$ where $r_\mathcal{B}$ is shown above as the quantile of the nearest-neighbor distances between base policy samples. Further, with this choice of radius, $r_\mathcal{B}$ becomes arbitrarily small as N increases, and the approximation converges uniformly over the action space in expectation:
\begin{equation}
\mathbb E_{\mathcal B}\!\left[
  \sup_{a\in A}
  \left|
    \ell_\mathrm{GSBR}(a,\mathcal B)
    -\ell_{\mathrm{supp}}(a)
  \right|
\right]
\longrightarrow 0
\qquad\text{as }N\to\infty.
\label{eq:tube-uniform-convergence}
\end{equation}
\end{theorem}
The full proof is in Appendix \ref{appendix-proof-approx-error}. The term $D^2(1-\lambda_r)^N$ accounts for the event that no reference
lies within radius $r$ of the in-distribution action nearest to $a$, and it vanishes as $N\to\infty$; The term $2Dr$ bounds the error caused by subtracting the quantile radius. 

We note that for generative policies, such as flow matching, they theoretically assign nonzero likelihood to every action \citep{lipman2023flowmatchinggenerativemodeling}. Instead, we can define our constraint with respect to some $\pi_\epsilon$ derived through filtering for actions such that $\pi_\beta(a|s) \ge \epsilon$, but we found this was not necessary in practice.

\subsection{Practical Implementation}

Algorithm~\ref{alg:tube} summarizes \ours{}. Training alternates between environment interaction and gradient
updates. Training always samples from a replay buffer $\mathcal{D}$ that mixes online and offline transitions. The learned actor receives the state and a set of reference actions $\mathcal{B}$ as inputs. The algorithm has hyperparameters: $N$ - number of sample actions, $q$ - quantile that affects the size of the radius, and $\alpha$ - that controls the strength of the behavior regularizer.

\begin{algorithm}[tbp]
\caption{Online RL with \ours{}}
\label{alg:tube}
\small
\begin{algorithmic}[1]
\Require $\pi_\beta$, $\mathcal{D}_{\mathrm{off}}$, $N$, $q$, $\alpha$.
\State Initialize $\pi_\theta$, $Q_{\phi}$, $Q_{\bar{\phi}}$,and $\mathcal{D}_{\mathrm{on}}$.
\For{each environment interaction}
    \State Sample $\mathcal{B}\sim\pi_\beta(\cdot\mid s)^N$;
    execute $a\sim\pi_\theta(\cdot\mid s,\mathcal{B})$ and store the transition in $\mathcal{D}_{\mathrm{on}}$.
    \For{each gradient update}
        \State Sample minibatch $(s,a,r,s') \sim
        \mathcal{D}_{\mathrm{off}}$ and $\mathcal{D}_{\mathrm{on}}$.
        \algphase{Train the critic}
        \State Sample $\mathcal{B}'\sim\pi_\beta(\cdot\mid s')^N$
        and $a'\sim\pi_\theta(\cdot\mid s',\mathcal{B}')$.
        \State Update $\phi$ to minimize
        $\mathbb{E}\!\left[\left(Q_{\phi}(s,a)-r-\gamma Q_{\bar\phi}(s',a')\right)^2\right]$.
        \algphase{Train the actor with geometric support regularization}
        \State Sample $\mathcal{B}=\{b_i\}_{i=1}^N\sim\pi_\beta(\cdot\mid s)^N$
        and $a_\theta\sim\pi_\theta(\cdot\mid s,\mathcal{B})$.
        \State $r_{\mathcal{B}}\gets\operatorname{Quantile}_q\!\left(
        \left\{\min_{j\ne i}\|b_i-b_j\|_2\right\}_{i=1}^N\right)$.
        \State Update $\theta$ to minimize
        $\mathbb{E}_{\mathrm{batch}}\!\left[-Q_\phi(s,a_\theta)
        +\alpha\left[\min_i\|a_\theta-b_i\|_2-r_{\mathcal{B}}\right]_+^2\right]$.
    \EndFor
\EndFor
\end{algorithmic}
\end{algorithm}

\section{Experimental Results}
In this section, we empirically evaluate the performance of \ours{} against previous online RL methods that leverage a frozen base policy across benchmarks covering both state and pixel inputs with MLP and VLA base policies. Throughout our experiments, we answer the following questions: (1) Is \ours{} an effective behavior regularizer for online RL, (2) how sensitive is \ours{} to the choice of support radius and number of samples, and (3) how well does \ours{} estimate the true support region?

\subsection{Experimental Setup}

\paragraph{Benchmarks. } We evaluate all methods on 19 tasks across four benchmarks, each having different properties that may change the distribution of base policies. In \textbf{RoboMimic}~\citep{mandlekar2022robomimic}, we select four challenging tasks: \texttt{can-mh}, \texttt{square-mh}, \texttt{transport-mh}, and \texttt{tool-hang-ph}. Each task contains demonstrations collected by human operators of various proficiency levels and captures challenges like precise alignment and bimanual coordination. We test five variants of the \texttt{cube-triple-task\{1-5\}} in \textbf{OGBench}~\citep{park2025ogbench}. These long-horizon tasks contain a play-style exploration dataset that may lead to a base policy with many action modes and test the RL actor's ability to drop bad actions while exploring better ones. The recently released benchmark \textbf{OCBench}~\citep{park2026behavioral} provides a more realistic dataset that emulates the statistical properties of human demonstrations. We select \texttt{block-lite-double-task\{1-2\}} and \texttt{chamber-lite-easy-task\{1-3\}}. In these three benchmarks, the base policy is an MLP behavior cloning policy trained on only the task-specific dataset, and both the base and RL policies receive state information from the simulator. To test each method on more realistic settings, we also experiment on \textbf{LIBERO}~\citep{liu2023libero} with the fine-tuned $\pi_{0.5}$-LIBERO base policy using pixel inputs. We hand-picked five tasks where the base VLA leaves room for improvement: \texttt{task\{14, 29, 54, 60, 64\}}. 
On these image-based tasks, the RL network trains a CNN from scratch. Since the $\pi_{0.5}$ policy is slow to sample from, we cache a set of reference actions and sample from these instead of sampling from the policy at each training step.

\paragraph{Baselines.}
We compare \ours{} with seven online RL baselines.
\textbf{TD3+BC}~\citep{fujimoto2021minimalist} trains a state-conditioned actor with a Q-maximization objective and a squared regression penalty to base policy samples. For expressive base policies, TD3+BC regularizes the action to the mean of the reference action distribution.
\textbf{RLT}~\citep{xu2026rlt} conditions a Gaussian chunk actor on a single sampled base policy chunk and penalizes deviation from that specific reference. While the conditioning alleviates the mode averaging problem, RLT cannot drop action modes. 
\textbf{Residual Policy}~\citep{silver2018residual,johannink2019residual} scales and adds a learned correction to a sampled base policy action, so its deviation from the base policy proposal is limited. We follow the formulation of~\citet{dong2026expoft} and learn an action chunked residual policy conditioned on the base policy input.
\textbf{BFN}~\citep{li2025qc} does not learn an explicit actor, instead formulating it as the base policy sample that maximizes the Q value. This is analogous to a hard support constraint since the implicit actor cannot generate actions outside the base policy's support.
\textbf{DSRL}~\citep{wagenmaker2025dsrl} learns to steer the frozen base policy through its latent noise rather than directly producing corrected actions, also limiting the possible actions to the support of the base policy. We use the DSRL-NA variant for the state-based tasks and the DSRL-SAC variant for LIBERO.
\textbf{EXPO-FT}~\citep{dong2026expoft} combines bounded residual edit policy with best-of-N sampling at test-time to achieve mode selection. For a fair comparison, we freeze the base policy. 
\textbf{DICE-RL}~\citep{sun2026dicerl} learns latent-conditioned residual corrections to base policy samples, applies selective BC regularization on the residual policy, and applies best-of-N sampling. 

All methods use the same offline data, online interaction budget, network size, and architecture. They also share training settings, including the action-chunk horizon, discount factor, and critic configuration. For each task, we evaluate each policy for 50 episodes and report the mean and standard error of the mean (SEM) across three seeds.

\begin{figure}[t]
    \centering
        
    \includegraphics[width=.8\linewidth]{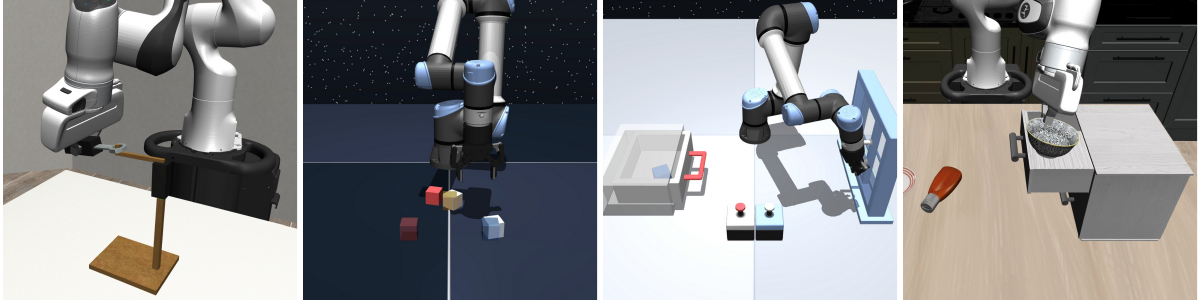}
    
    \caption{\textbf{Benchmarks}. We evaluate our method across the four benchmarks above. From left to right, the tasks are RoboMimic \texttt{tool-hang}, OGBench \texttt{cube-triple-task4}, OCBench \texttt{chamber-lite-easy-task3}, and LIBERO \texttt{task29}.}
    
    \label{fig:placeholder}
\end{figure}

\begin{figure}[t]
    \centering
    \includegraphics[width=1\linewidth]{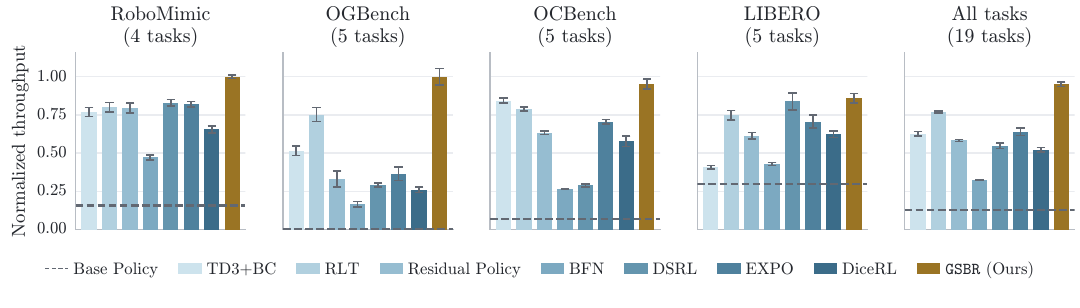}
    
    \caption{\textbf{Normalized throughput across benchmarks}. \ours{} achieves the highest average normalized throughput across all 19 tasks and leads all baselines on \texttt{RoboMimic}, \texttt{OGBench}, and \texttt{OCBench}. Throughput (success rate / mean episode length) is normalized by the best method's seed-average value per task, then averaged across tasks. Error bars show SEM across seeds.}
    \label{fig:results}
\end{figure}

\begin{table*}[ht!]
    \centering
\vspace{1em}
    \footnotesize
    \setlength{\tabcolsep}{7pt}
    \begin{tabular*}{\textwidth}{@{\extracolsep{\fill}}lccccc@{}}
        \toprule
        Method & RoboMimic & OGBench & OCBench & LIBERO & All tasks \\
        & (4 tasks) & (5 tasks) & (5 tasks) & (5 tasks) & (19 tasks) \\
        \midrule
        Base Policy & $44$ & $0$ & $41$ & $49$ & $33$ \\
        \midrule
        TD3+BC & $96 \pm 2$ & $73 \pm 1$ & $98 \pm 0$ & $54 \pm 2$ & $79 \pm 1$ \\
        RLT & $96 \pm 2$ & $90 \pm 5$ & $98 \pm 0$ & $\mathbf{86} \pm 2$ & $92 \pm 1$ \\
        Residual Policy & $94 \pm 2$ & $68 \pm 10$ & $95 \pm 0$ & $78 \pm 1$ & $83 \pm 2$ \\
        BFN & $86 \pm 2$ & $69 \pm 5$ & $88 \pm 2$ & $66 \pm 1$ & $77 \pm 1$ \\
        DSRL & $97 \pm 1$ & $67 \pm 5$ & $87 \pm 2$ & $81 \pm 2$ & $82 \pm 2$ \\
        EXPO & $97 \pm 1$ & $80 \pm 1$ & $98 \pm 1$ & $82 \pm 2$ & $89 \pm 1$ \\
        DiceRL & $95 \pm 1$ & $40 \pm 1$ & $97 \pm 1$ & $74 \pm 1$ & $75 \pm 1$ \\
        \midrule
        \ours{} (Ours) & $\mathbf{98} \pm 0$ & $\mathbf{99} \pm 1$ & $\mathbf{99} \pm 0$ & $85 \pm 1$ & $\mathbf{95} \pm 0$ \\
        \bottomrule
    \end{tabular*}
    \caption{\textbf{Average success rate (\%) across benchmarks}. Uncertainty is SEM across three seed-wise task averages. Values are rounded to the nearest integer; bold marks the highest unrounded mean in each column. Base Policy SEM is omitted because complete per-seed evidence is unavailable.}
    
    \label{tab:benchmark_success_rates}
\end{table*}

\subsection{Results}

\paragraph{Is \ours{} an effective behavior regularizer for online RL?}

\ours{} demonstrates strong empirical performance across benchmarks compared to the baselines. As shown in Figure~\ref{fig:results} and Table~\ref{tab:benchmark_success_rates}, across 19 tasks, \ours{} achieves 95\% success rate and 0.95 normalized throughput, while the next-best baseline, RLT, achieves 92\% success rate and 0.77 normalized throughput. On RoboMimic, while most methods have a similar success rate at around 95\%, \ours{} achieves the highest throughput, surpassing the next-best method, DSRL, by 20\%. On OGBench, \ours{} achieves a near-perfect success rate across all five tasks, which is about 9 percentage points higher in success rate and 33\% (relative) higher throughput compared to RLT, the strongest baseline. On OCBench, most methods achieve a near-perfect success rate yet \ours{} has approximately 13\% higher throughput than TD3+BC. 

Further, we evaluate the performance of our behavior regularizer on LIBERO with a modern VLA as the base policy. Using $\pi_{0.5}$-LIBERO, the strongest baseline is DSRL in terms of throughput at 0.84 and RLT in terms of success rate at 86\%. \ours{} achieves $85\%$ success rate and $ 0.86$ normalized throughput, closely matching or beating the best baselines on either metric. 
These results demonstrate that \ours{} provides an effective behavior regularizer that empirically outperforms baselines in success rate and throughput, indicating that our regularizer allows the maximum deviation from the base policy while providing stabilizing guidance. For additional experimental detail and results, refer to Appendix \ref{app:state} and \ref{apdx:libero-stuff}




\begin{wraptable}{r}{0.48\textwidth}
  \noindent\hfill\begin{minipage}{0.9\linewidth}
  \centering
  \small
  \captionsetup{justification=raggedright}
  \renewcommand{\arraystretch}{1.18}
  \setlength{\tabcolsep}{4pt}
  \setlength{\abovecaptionskip}{5pt}
  \setlength{\belowcaptionskip}{0pt}
  \begin{tabular*}{\linewidth}{@{\extracolsep{\fill}}lc@{}}
    \toprule
    Reference count $N$ & Normalized TP \\
    \midrule
    8 (Ours) & $\textbf{1.00} \pm \textbf{0.05}$ \\
    16 & $0.91 \pm 0.02$ \\
    32 & $0.93 \pm 0.02$ \\
    \bottomrule
  \end{tabular*}
  \caption{\textbf{Reference-count ablation.} On \texttt{cube-triple}, the number of base policy samples, $N$, has little effect on performance.}
  \label{tab:cube-triple-tube-n}
  \vspace{2em}

  \par\medskip
  \begin{tabular*}{\linewidth}{@{\extracolsep{\fill}}lc@{}}
    \toprule
    Support radius ablation & Normalized TP \\
    \midrule
    $r_{\mathcal{B}}=0$ & $0.66 \pm 0.07$ \\
    \ours{} (Ours) & $\textbf{1.00} \pm \textbf{0.05}$ \\
    \bottomrule
  \end{tabular*}
  \caption{\textbf{Support-radius ablation.} Removing the support radius, $r_{\mathcal{B}}$, significantly reduces normalized throughput on \texttt{cube-triple}.}
  \label{tab:cube-triple-radius}
  \end{minipage}
\end{wraptable}
\paragraph{How sensitive is \ours{} to the choice of support radius and number of samples?}

We sweep over a range of $N$, the number of base policy samples used to compute the behavior regularizer loss on \texttt{cube-triple} (Table~\ref{tab:cube-triple-tube-n}) and find little effect on the performance of \ours{}. The normalized throughput for $N=8$, $N=16$, and $N=32$ is 1.0, 0.91, and 0.93, respectively. These results suggest that as few as eight samples from the base policy suffice for effective behavior regularization on these tasks. 

We also ablate the effect of computing the support radius $r_{\mathcal{B}}$ (Table~\ref{tab:cube-triple-radius}). With $r_{\mathcal{B}}=0$, the throughput degrades significantly to only 0.66. This ablation supports the benefit of leaving neighborhoods around reference actions unpenalized, unlike in other prior works, allowing the actor to refine its actions within these neighborhoods without being pulled toward individual samples.

\par
\begin{figure}
    \centering
    \includegraphics[width=1\linewidth]{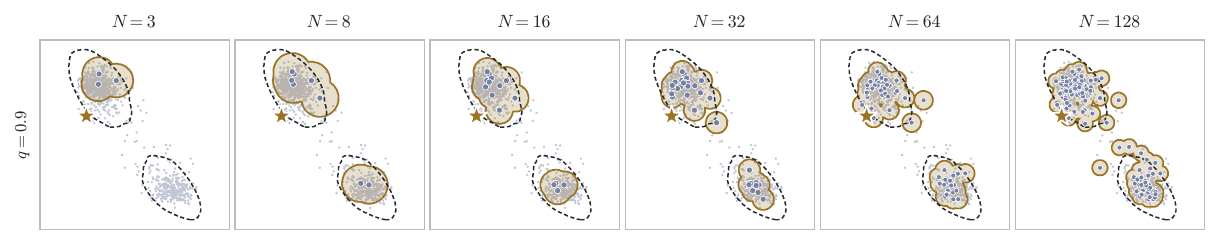}
    
    \caption{\textbf{Visualization of the true and estimated support region on \texttt{cube-triple}}. The estimated support region (in bronze) closely matches the true support using only 8 samples. The RL actor learned an action just outside the support region (marked with star).}
    
    \label{fig:interp}
\end{figure}

\WFclear
\paragraph{How well do we approximate the true support region?}
Figure~\ref{fig:interp} shows the empirical support of the base policy visualized on a bimodal state in \texttt{cube-triple}. We visualize the first two action dimensions using $q=0.9$, and the zero-penalty estimated support region $C$ is marked in golden bronze. We estimate the true support of the base policy with KDE cutoff at 95\% from 5,000 base-policy samples. With only $N=3$ references, the sampled actions cover one mode and miss the other. At $N=8$, the zero-penalty region already captures both modes while leaving the low-density gap between them penalized. As $N$ increases, the support radius shrinks as the local spacing between samples becomes tighter. On this particular state, the trained RL actor predicts an action slightly beyond the true support, which incurs little loss as it is close to the support region.

\section{Conclusions}
Foundation models offer an appealing behavior prior for sample-efficient
online RL, which raises the question of how to regularize an actor toward such a prior while still allowing it to improve.
We introduce \ours{}, a geometric support behavior regularizer that provides sufficient guidance without overconstraining policy improvement. We provide a theoretically motivated implementation that estimates the ideal loss using base policy samples. \ours{} demonstrated strong empirical performance: Across 19 tasks on four robotic benchmarks, \ours{} raises the average
success rate from 33\% for the base policy to 95\%, achieving the highest
average success rate and throughput among all compared methods, with
about 23\% higher normalized throughput than the best prior method.

One limitation of the method is that it uses multiple base policy samples at test-time, which may impose compute overhead and introduce more inference latency. Further, \ours{} introduces an extra hyperparameter, the quantile $q$, compared to other common methods. Although we tuned $q$ for the best performance, fortunately, Figure~\ref{fig:sensitivity} shows reasonable performance across the full range of possible values. Validating \ours{} on real robots is a natural next step. Future work could also explore alternative ways to construct the support radius that could improve performance or reduce hyperparameter count.

\section*{Acknowledgments}
This research was supported by ONR N00014-25-1-2060. We would like to thank Seohong Park, Louis McConnell, Chongyi Zheng, and Zhiyuan Zhou for helpful discussion and feedback on the paper; we thank Jagdeep Bhatia and Qiyang Li for providing code prototype and early discussion. We thank the Google TPU Research Cloud (TRC) program for providing TPU compute resources. This research also used the Savio computational cluster resource provided by the Berkeley Research Computing program at UC Berkeley. 

\section*{Author Contributions}
CX led the project, developed the core idea and method, designed and conducted experiments, and led manuscript writing and figure preparation. JB developed the theory and experimental analysis, helped iterate the method, contributed to the training infrastructure, conducted experiments, and contributed to writing. LK and SL provided valuable guidance and feedback throughout the project and contributed to manuscript writing and revision.

\bibliography{iclr2027_conference}
\bibliographystyle{iclr2027_conference}

\appendix
\makeatletter
\setlength{\@fptop}{0pt}
\setlength{\@fpsep}{12pt}
\makeatother
\section{Appendix}

\subsection{Practical Implementation}
\label{sec:online-learning}

In this section, we discuss some of the practical implementation choices for our method.

\textbf{Action chunking.} Both the base policy and RL policy learn over action chunks of $H$ consecutive steps, which vary between three and ten depending on the domain. During rollout and evaluation, we execute the entire action chunk before querying the next one.

\textbf{Reference action conditioning.} Following~\citet{xu2026rlt}, we condition the RL actor on $N$ reference action samples to provide additional learning signal and guidance. During training, we apply 50\% dropout to prevent the actor from copying the reference action. 

\textbf{Using demo data.} Following~\citet{ball2023rlpd}, we store separate offline and online replay buffers $\mathcal{D}_{\mathrm{off}}$ and $\mathcal{D}_{\mathrm{on}}$ to maintain a 50/50 sampling ratio between demo and online data.

\textbf{Critic ensemble.} We train an ensemble of 10 Q networks $\{Q_{\phi_k}\}_{k=1}^{K}$ and use the following pessimistic target value
\begin{equation}
y= R + \gamma^H m\left(\mu_{\bar\phi}(s',\widetilde{a}')-\tfrac{1}{2}\sigma_{\bar\phi}(s',\widetilde{a}')\right)
\label{eq:td_target}
\end{equation}
where $\mu_{\bar\phi}(s',\widetilde{a}')$ and $\sigma_{\bar\phi}(s',\widetilde{a}')$ are the mean and standard deviation over the 10 target Q networks. Here $R$ is the discounted chunk reward (Equation~\ref{eq:chunk-reward}), $m$ is the bootstrap mask, and $\widetilde{a}'\sim\pi_\theta(\cdot\mid s',\mathcal{B}')$ with next-state references $\mathcal{B}'\sim\pi_\beta(\cdot\mid s')^N$. The chunk discount is $\gamma^H$ for a per-step discount; for LIBERO, we use the configured per-chunk discount directly. The main-text critic objective and Algorithm~\ref{alg:tube} use single-step notation.

\subsection{Proof of Theorem \ref{thm:tube-error}}
\label{appendix-proof-approx-error}

\paragraph{Theorem \ref{thm:tube-error} restated:}
\begingroup
\itshape

Assume the action space $A$ is compact with diameter \(D\), and that $\pi_\beta(a \mid s)$ has continuous density.
Fix any $a \in A$, and define our reference action set $\mathcal{B}$ as above.

Then, for any radius $r>0$, define a $\lambda_r > 0$ as:
\begin{equation*}
\lambda_r := \inf_{c\in S_\beta} \Pr_{b\sim \pi_\beta}\bigl(\|b-c\|_2\le r\bigr)
\end{equation*}representing the worst-case probability over our support set that a sampled action from $\pi_\beta$ falls within distance $r$ of a given action in $S_\beta$.

The following bounds then hold: 
\begin{align*}
\ell_{\mathrm{supp}}(a)
&\le \ell_\mathrm{GSBR}(a,\mathcal B)+2Dr,\\
\mathbb E_\mathcal{B}\!\left[
  \left|\ell_\mathrm{GSBR}(a,\mathcal B)
       -\ell_{\mathrm{supp}}(a)\right|
\right]
&\le 2Dr+D^2(1-\lambda_r)^N.
\end{align*}
The upper bound in Equation~\ref{eq:tube-one-sided}
also holds with $r=r_\mathcal{B}$ where $r_\mathcal{B}$ is  shown above as the quantile of the nearest-neighbor distances between base policy samples. Further, with this choice of radius, $r_\mathcal{B}$ becomes arbitrarily small as N increases, and the approximation converges uniformly over the action space in expectation:
\begin{equation*}
\mathbb E_{\mathcal B}\!\left[
  \sup_{a\in A}
  \left|
    \ell_\mathrm{GSBR}(a,\mathcal B)
    -\ell_{\mathrm{supp}}(a)
  \right|
\right]
\longrightarrow 0
\qquad\text{as }N\to\infty.
\end{equation*}
\par
\endgroup

\begin{proof}
We first show that $\lambda_r>0$. Since $S_\beta$ is compact,
it can be covered by finitely many balls of radius $r/2$
centered in $S_\beta$. Each has positive probability under $\pi_\beta$,
because its center belongs to the support of $\pi_\beta$.
For every $c\in S_\beta$, some covering ball has its center
within $r/2$ of $c$, so that entire ball lies within
distance $r$ of $c$. Thus $\lambda_r$ is at least the
smallest probability among these finitely many covering
balls, which is positive.

Now fix an action $a\in \mathcal A$ and write
\[
\rho_a:=d(a,S_\beta),
\qquad
X:=d(a, \mathcal B),
\qquad
t:=[X-r]_+.
\]
Because $\rho_a=d(a,S_\beta)$ and $B\subseteq S_\beta$,
we have $\rho_a\le X$.
Since $\rho_a\le X$, $X-t=\min\{X,r\}\le r$, and
$X+t\le2D$, we have
\[
\ell_{\mathrm{supp}}(a)-\ell_\mathrm{GSBR}(a,\mathcal B)
=\rho_a^2-t^2
\le X^2-t^2
=(X-t)(X+t)
\le2Dr.
\]
Thus, adding $\ell_{\ours}(a, \mathcal B)$ to both sides shows that the \ours{} penalty plus $2Dr$ upper bounds the support
penalty:
\[
\ell_{\mathrm{supp}}(a)
\le \ell_\mathrm{GSBR}(a, \mathcal B)+2Dr.
\]

Next for the error bound, since $S_\beta$ is compact, there is a closest point
$b^\star\in S_\beta$ with $\|a-b^\star\|_2=\rho_a$.
A reference within distance $r$ of $b^\star$ is also within
distance $\rho_a+r$ of $a$, by the triangle inequality.
Thus,
\[
\Pr_{b\sim\pi_\beta}
\bigl(\|b-b^\star\|_2\le r\bigr)
\le
\Pr_{b\sim\pi_\beta}
\bigl(\|b-a\|_2\le\rho_a+r\bigr)
\]
And using the definition of $\lambda_r$ as the lower bound,  $\lambda_r \leq \Pr_{b\sim\pi_\beta} \bigl(\|b-a\|_2\le\rho_a+r\bigr) $

Because the references are independent of one another and
of $a$, the probability that all $N$ references miss this
neighborhood is at most
\[
\Pr_{\mathcal B}(X>\rho_a+r\mid a)\le(1-\lambda_r)^N.
\]

We now consider a decomposition of the expectation by partitioning on the event that $X\le\rho_a+r$. 

On the event $X\le\rho_a+r$, subtracting $r$ from both sides gives
$X-r\le\rho_a$, and since $\rho_a \geq 0$, $t=[X-r]_+\le\rho_a$. Therefore, the support penalty
is at least the \ours{} penalty, so the bound from above gives
\[
\left|
\ell_\mathrm{GSBR}(a,\mathcal B)-\ell_{\mathrm{supp}}(a)
\right|
=\rho_a^2-t^2
\le2Dr.
\]

On the complementary event $X>\rho_a+r$, both penalties
lie in $[0,D^2]$, so their absolute difference is at most
$D^2$. Combining the two events gives
\[
\begin{aligned}
\E_B\!\left[
\left|
\ell_\mathrm{GSBR}(a,\mathcal B)-\ell_{\mathrm{supp}}(a)
\right|
\,
\right]
&\le
2Dr\,\Pr(X\le\rho_a+r\mid a)
+D^2\,\Pr(X>\rho_a+r\mid a)\\
&\le 2Dr+D^2(1-\lambda_r)^N.
\end{aligned}
\]

Finally, consider the quantile radius $r_\mathcal{B}$.
The one-sided bound above is pointwise in $r$ and
$\mathcal B$, so it also holds with $r=r_\mathcal{B}$.

Define the largest distance from the support to the reference
set by
\[
h_N:=\sup_{c\in S_\beta}d(c,\mathcal B).
\]
For any $a\in A$, using the notation above with $r=r_\mathcal{B}$,
\[
\rho_a\le X\le\rho_a+h_N,
\qquad
\rho_a-r_\mathcal{B}\le t=[X-r_\mathcal{B}]_+\le\rho_a+h_N.
\]
Thus $|t-\rho_a|\le\max\{h_N,r_\mathcal{B}\}$. Since both $t$ and
$\rho_a$ lie in $[0,D]$, we have, for every $a\in A$,
\[
\left|\ell_\mathrm{GSBR}(a,\mathcal B)
-\ell_{\mathrm{supp}}(a)\right|
=|t^2-\rho_a^2|
\le2D\max\{h_N,r_\mathcal{B}\}.
\]

To show that this error vanishes, fix any $\delta>0$.
Cover $S_\beta$ by $M$ closed balls of radius $\delta/2$
centered in $S_\beta$. Each ball has probability at least
$\lambda_{\delta/2}>0$. If every covering ball contains a
reference, every support point is within distance $\delta$
of a reference. Therefore,
\[
\Pr_{\mathcal B}(h_N>\delta)
\le M(1-\lambda_{\delta/2})^N.
\]

The same cover controls the quantile radius. If $u_i>\delta$,
any covering ball containing $b_i$ cannot contain another
reference, since the distance between two points in that ball
is at most $\delta$. Hence at most $M$ references have
$u_i>\delta$. For all sufficiently large $N$,
$N-M\ge\lceil qN\rceil$, so $r_\mathcal{B}\le\delta$.

For all sufficiently large $N$, we have $r_\mathcal{B}\le\delta$.
On the event $h_N\le\delta$, the absolute penalty error
is therefore at most $2D\delta$ simultaneously for every
$a\in \mathcal A$. On the complementary event, it is at most $D^2$.
Taking the supremum over actions and then the expectation
over reference samples gives
\begin{equation}
\begin{aligned}
\mathbb E_{\mathcal B}\!\left[
  \sup_{a\in A}
  \left|
    \ell_\mathrm{GSBR}(a,\mathcal B)
    -\ell_{\mathrm{supp}}(a)
  \right|
\right]
&\le 2D\delta+D^2\Pr_{\mathcal B}(h_N>\delta)\\
&\le 2D\delta+D^2M(1-\lambda_{\delta/2})^N.
\end{aligned}
\end{equation}

Let $\eta>0$ be any desired error tolerance. Choose $\delta>0$
small enough that $2D\delta<\eta/2$. For this fixed $\delta$,
$D^2M(1-\lambda_{\delta/2})^N \rightarrow 0$ as $N \rightarrow \infty$, thus it must be less than  $\eta/2$ for all
sufficiently large $N$. The expected uniform error is therefore
less than $\eta$ for all sufficiently large $N$. Since $\eta>0$
is arbitrary, the expected uniform error converges to zero
as $N\to\infty$, proving Equation~\ref{eq:tube-uniform-convergence}.\end{proof}

\subsection{Implementation Choices}
\label{sec:implementation-choices}

\paragraph{Action chunking.} For robotic control, we instantiate the action space using chunks
of $H$ consecutive control actions. Each action chunk is flattened into
a vector of dimension $d=Hd_a$, where $d_a$ is the dimension of
an individual control action. In this implementation, $a$, $b$,
and the references in $\mathcal{B}$ denote action chunks.
Training uses $H$-step replay transitions $(s,a,R,s')$,
where $a$ concatenates the recorded control actions
$a_t,\ldots,a_{t+H-1}$, $s = s_t$, $s'=s_{t+H}$ and
\begin{equation}
R=\sum_{h=0}^{H-1}\gamma^h r(s_{t+h},a_{t+h}),
\qquad s=s_t,\quad s'=s_{t+H}.
\label{eq:chunk-reward}
\end{equation}

\paragraph{Reference action conditioning.} We parameterize the actor as a tanh-squashed Gaussian policy conditioned on $s$. To reduce the burden on the actor on learning the full distribution of the base policy and provide early-training exploration signal, we adopt the reference action passthrough mechanism from \citet{xu2026rlt} and additionally condition the actor on all sampled reference actions $\mathcal{B}$. During training, we zero the entire reference input with probability $p_{\mathrm{drop}}$. 
In early testing, we found that this mechanism improves initial performance without having to sample from the base policy during deployment. 

\paragraph{Critic ensembles.} To stabilize Q-learning, we train an ensemble of $K=10$
Q-functions $\{Q_{\phi_k}\}_{k=1}^{K}$.
We compute the mean and standard deviation of their target
predictions as
\begin{equation}
\mu_{\bar\phi}(s',a')
= \frac{1}{K}\sum_{k=1}^{K}Q_{\bar\phi_k}(s',a'), \quad
\sigma_{\bar\phi}(s',a')
= \sqrt{\frac{1}{K}\sum_{k=1}^{K}
\left(Q_{\bar\phi_k}(s',a')-\mu_{\bar\phi}(s',a')\right)^2}.
\end{equation}
For TD learning, we use the pessimistic target and the critic loss is
\begin{equation}
\mathcal{L}_{Q}(\phi_k)
=\mathbb{E}_{\substack{
(s,a,R,s')\sim\mathcal{D}\\
\mathcal{B}'\sim\pi_\beta(\cdot\mid s')^N\\
a'\sim\pi_\theta(\cdot\mid s',\mathcal{B}')
}}
\left[
\left[Q_{\phi_k}(s,a)-R-\gamma^H m\left(\mu_{\bar\phi}(s',a')
-\tfrac{1}{2}\sigma_{\bar\phi}(s',a')\right)\right]^2
\right].
\label{eq:critic-objective}
\end{equation}
The discount convention follows Equation~\ref{eq:td_target}: for LIBERO, the configured per-chunk discount replaces $\gamma^H$.
For the actor update, we simply take the mean of the Q-values.

\paragraph{Buffer sampling ratio.} We maintain separate offline and online replay buffers,
$\mathcal{D}_{\mathrm{off}}$ and $\mathcal{D}_{\mathrm{on}}$.
Following the symmetric replay sampling strategy of
RLPD~\citep{ball2023rlpd}, each training minibatch contains
equal numbers of $H$-step sequences from the two buffers.
This retains access to prior experience as new interactions
accumulate.

\paragraph{Rollouts and deployment.} During a rollout, we sample reference chunks and obtain
an action chunk from the actor without reference dropout.
During deployment, we can optionally skip querying the base policy and run the actor with reference actions set to zero.


\subsection{Training details}
\label{appdx:training-details}

We describe the shared training protocol for the state-based benchmarks,
followed by method-specific settings and the configuration used for LIBERO.
Unless specified otherwise, the state-based methods use the protocol below. Our training and evaluations were done on either A5000 GPUs or TPU V5E chips. Common training settings are described in Table~\ref{tab:training-settings}. We also report the hyperparameter sweep range and the selected values of each method in Table~\ref{tab:gsbr-search} and Table~\ref{tab:baseline-search}. We select the hyperparameter value with priority on average success rate, and favor the configuration with higher average throughput if multiple runs have success rate within 2\% of each other. 

\subsubsection{State-based benchmarks}
\label{app:state}

\paragraph{Base policies and datasets.}
For each training seed, we use a seed-matched flow-matching behavior cloning
policy trained offline for $10^6$ training steps. The base policy produces
action chunks and remains frozen during online RL. On RoboMimic and OGBench
Cube Triple, it uses a four-layer MLP with 512 units per layer and 16 Euler
integration steps. On the OCBench Block and Chamber tasks, it uses a residual
MLP with four 1024-unit layers and 10 Euler integration steps. We clip sampled base-policy actions to the action bounds.

For RoboMimic, we use the low-dimensional v1.5 datasets: multi-human (MH)
demonstrations for Can, Square, and Transport, and proficient-human (PH)
demonstrations for Tool Hang due to it not providing a MH dataset. For OGBench, we use the cube triple play dataset. For OCBench, we use the type 51 dataset shards and normalize each action dimension using its 1st and 99th percentiles.

\paragraph{Replay and online training.}
The RL components receive no offline pretraining. We initialize online replay
with 50 episodes collected by the base policy, then train with minibatches of
256 sequences, split equally between the offline dataset and online replay.
We use two gradient updates per environment step, except for the DICE-RL
where we take the max performance between a UTD 2 and the effective UTD used for their results in robosuite. We do not filter transitions by success
and use the environment rewards. RoboMimic supplies a binary success reward,
and episodes terminate on success; Cube Triple uses the OGBench reward.


\subsubsection{LIBERO}
\label{apdx:libero-stuff}
\paragraph{Tasks and observations.}
We evaluate on LIBERO-90 tasks 14, 29, 54, 60, and 64, with an episode limit of
400 environment steps and action chunks of length $H=10$. The frozen base
policy is $\pi_{0.5}$ with the \texttt{pi05\_libero} configuration, which
receives images from both cameras and proprioception. The RL actor and critic
use CNNs with a $64\times64$ image and an eight-dimensional state as input.

\paragraph{Training and evaluation.}
LIBERO training uses online data only, without demonstrations. Base-policy
reference chunks are cached and reused. Each unsuccessful chunk receives
reward $-1$, and a successful chunk receives reward $0$. The discount factor
is $0.975$ per chunk, and TD targets are not clipped. We train for 1,000
trajectories with 20 gradient updates per action chunk. 

\paragraph{Method-specific settings.}
For \ours{}, we draw $N=8$ reference chunks from the cache and do not apply
reference dropout. We use a separate radius for each of the seven action
dimensions due to the chunk horizon, with each radius shared across the chunk's time steps. 

\paragraph{LIBERO regularizer.}
For chunks with $H=10$ and $d_a=7$, we sample $N=8$ references
$\mathcal B=\{b_i\}_{i=1}^N$. Define
\begin{align}
\rho_k(x,y)
&:= \left(\frac{1}{H}\sum_{t=1}^H
(x_{t,k}-y_{t,k})^2\right)^{1/2},\\
j_i &\in \operatorname*{arg\,min}_{j\ne i}\|b_i-b_j\|_F^2,
\qquad
r_{\mathcal B,k}
:= \operatorname{Quantile}_q
\bigl(\{\rho_k(b_i,b_{j_i})\}_{i=1}^N\bigr),\\
\ell_{\mathrm{GSBR}}^{\mathrm{LIBERO}}(a,\mathcal B)
&:= \min_i \frac{1}{d_a}\sum_{k=1}^{d_a}
[\rho_k(a,b_i)-r_{\mathcal B,k}]_+^2.
\end{align}
We compute quantiles using linear interpolation.
Nearest neighbors are selected using full chunks, while
each action dimension has its own radius. The minimum selects
one reference for all dimensions; the penalty vanishes when
all dimension-wise distances to that reference lie within
their radii. We use this loss in the actor objective and sweep
$\alpha$ and $q$.

\subsection{Full results}
\begin{figure}
    \centering
    \includegraphics[width=1\linewidth]{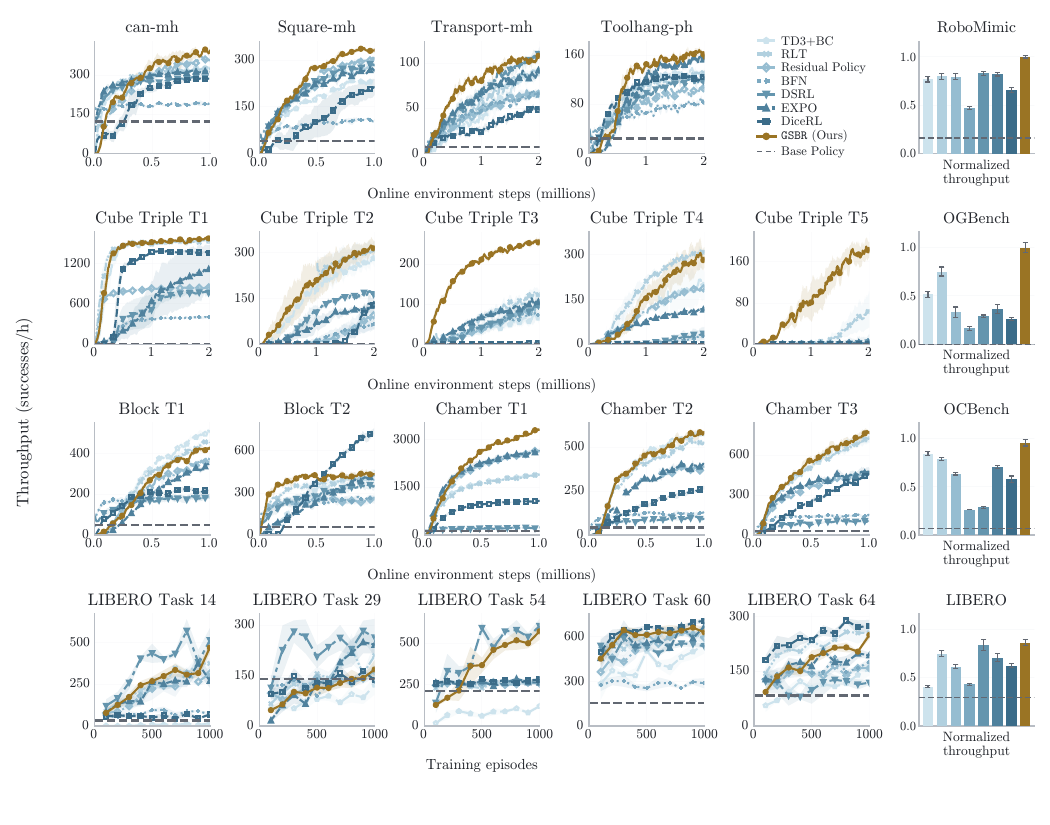}
    
    \caption{\textbf{Absolute throughput (successes/h) over training steps}}
    \label{fig:full-results-apdx}
\end{figure}

Our full per task results are shown in table \ref{tab:results-sr}, and \ref{tab:results-tp}, and visualized in figure \ref{fig:full-results-apdx}.

\label{app:additional_results}
\begin{table*}
\centering
\scriptsize
\setlength{\tabcolsep}{2.5pt}
\renewcommand{\arraystretch}{1.15}
\begin{tabular}{@{}lc|ccccccc>{\columncolor{GSBRBronze!7}}c@{}}
\toprule
Task & BC & TD3+BC & RLT & \shortstack{Residual policy} & BFN & DSRL & EXPO-FT & DICE-RL & \ours{} (ours) \\
\midrule
\multicolumn{10}{@{}l}{\textit{RoboMimic}} \\[2pt]
can-mh & $73{\pm}1$ & $\mathbf{99}{\pm}0$ & $\mathbf{99}{\pm}0$ & $97{\pm}1$ & $94{\pm}2$ & $\mathbf{99}{\pm}0$ & $\mathbf{99}{\pm}0$ & $\mathbf{99}{\pm}0$ & $98{\pm}1$ \\
square-mh & $37{\pm}1$ & $96{\pm}1$ & $96{\pm}1$ & $98{\pm}0$ & $78{\pm}3$ & $98{\pm}0$ & $98{\pm}0$ & $96{\pm}2$ & $\mathbf{99}{\pm}0$ \\
transport-mh & $15{\pm}2$ & $88{\pm}7$ & $90{\pm}6$ & $83{\pm}7$ & $81{\pm}4$ & $92{\pm}2$ & $93{\pm}1$ & $84{\pm}3$ & $\mathbf{96}{\pm}1$ \\
tool-hang-ph & $50{\pm}2$ & $97{\pm}0$ & $97{\pm}0$ & $97{\pm}2$ & $90{\pm}2$ & $96{\pm}2$ & $95{\pm}1$ & $\mathbf{98}{\pm}0$ & $98{\pm}0$ \\
\midrule
\multicolumn{10}{@{}l}{\textit{OGBench cube-triple}} \\[2pt]
Task 1 & $0{\pm}0$ & $\mathbf{100}{\pm}0$ & $\mathbf{100}{\pm}0$ & $\mathbf{100}{\pm}0$ & $\mathbf{100}{\pm}0$ & $\mathbf{100}{\pm}0$ & $98{\pm}1$ & $\mathbf{100}{\pm}0$ & $\mathbf{100}{\pm}0$ \\
Task 2 & $0{\pm}0$ & $98{\pm}0$ & $\mathbf{99}{\pm}0$ & $62{\pm}31$ & $80{\pm}9$ & $98{\pm}0$ & $96{\pm}2$ & $89{\pm}5$ & $\mathbf{99}{\pm}0$ \\
Task 3 & $0{\pm}0$ & $72{\pm}7$ & $92{\pm}1$ & $80{\pm}16$ & $82{\pm}4$ & $85{\pm}1$ & $96{\pm}2$ & $6{\pm}3$ & $\mathbf{100}{\pm}0$ \\
Task 4 & $0{\pm}0$ & $91{\pm}5$ & $\mathbf{100}{\pm}0$ & $98{\pm}1$ & $66{\pm}5$ & $46{\pm}23$ & $97{\pm}1$ & $1{\pm}1$ & $98{\pm}1$ \\
Task 5 & $0{\pm}0$ & $0{\pm}0$ & $58{\pm}29$ & $0{\pm}0$ & $15{\pm}15$ & $4{\pm}2$ & $8{\pm}4$ & $0{\pm}0$ & $\mathbf{95}{\pm}1$ \\
\midrule
\multicolumn{10}{@{}l}{\textit{OCBench Lite}} \\[2pt]
Block 1 & $33{\pm}5$ & $95{\pm}1$ & $95{\pm}1$ & $91{\pm}1$ & $92{\pm}1$ & $89{\pm}1$ & $\mathbf{97}{\pm}1$ & $93{\pm}2$ & $\mathbf{97}{\pm}1$ \\
Block 2 & $36{\pm}2$ & $\mathbf{99}{\pm}0$ & $98{\pm}0$ & $90{\pm}4$ & $90{\pm}2$ & $95{\pm}2$ & $94{\pm}1$ & $98{\pm}0$ & $98{\pm}0$ \\
Chamber 1 & $72{\pm}3$ & $\mathbf{100}{\pm}0$ & $\mathbf{100}{\pm}0$ & $\mathbf{100}{\pm}0$ & $97{\pm}0$ & $96{\pm}2$ & $\mathbf{100}{\pm}0$ & $\mathbf{100}{\pm}0$ & $\mathbf{100}{\pm}0$ \\
Chamber 2 & $36{\pm}5$ & $\mathbf{99}{\pm}0$ & $96{\pm}2$ & $96{\pm}2$ & $76{\pm}4$ & $72{\pm}7$ & $98{\pm}1$ & $94{\pm}1$ & $98{\pm}1$ \\
Chamber 3 & $26{\pm}3$ & $98{\pm}1$ & $98{\pm}1$ & $97{\pm}1$ & $84{\pm}5$ & $80{\pm}4$ & $98{\pm}1$ & $99{\pm}0$ & $\mathbf{100}{\pm}0$ \\
\midrule
\multicolumn{10}{@{}l}{\textit{LIBERO}} \\[2pt]
Task 14 & $16$ & $8{\pm}4$ & $80{\pm}2$ & $74{\pm}4$ & $30{\pm}4$ & $\mathbf{87}{\pm}2$ & $70{\pm}5$ & $30{\pm}2$ & $86{\pm}2$ \\
Task 29 & $52$ & $46{\pm}8$ & $65{\pm}7$ & $59{\pm}5$ & $49{\pm}2$ & $70{\pm}7$ & $\mathbf{73}{\pm}4$ & $51{\pm}4$ & $60{\pm}6$ \\
Task 54 & $84$ & $56{\pm}3$ & $95{\pm}1$ & $93{\pm}0$ & $96{\pm}1$ & $96{\pm}2$ & $94{\pm}1$ & $\mathbf{97}{\pm}0$ & $94{\pm}2$ \\
Task 60 & $54$ & $94{\pm}3$ & $98{\pm}0$ & $96{\pm}3$ & $80{\pm}4$ & $\mathbf{100}{\pm}0$ & $98{\pm}0$ & $\mathbf{100}{\pm}0$ & $98{\pm}1$ \\
Task 64 & $40$ & $66{\pm}6$ & $87{\pm}3$ & $66{\pm}3$ & $70{\pm}2$ & $51{\pm}2$ & $76{\pm}2$ & $\mathbf{90}{\pm}3$ & $84{\pm}2$ \\
\bottomrule
\end{tabular}
\caption{\textbf{Final success rate (\%) across all tasks}. Values are mean $\pm$ SEM.}
\label{tab:results-sr}
\end{table*}

\begin{table*}
\centering
\scriptsize
\setlength{\tabcolsep}{2.5pt}
\renewcommand{\arraystretch}{1.15}
\begin{tabular}{@{}lc|ccccccc>{\columncolor{GSBRBronze!7}}c@{}}
\toprule
Task & BC & TD3+BC & RLT & \shortstack{Residual policy} & BFN & DSRL & EXPO-FT & DICE-RL & \ours{} (ours) \\
\midrule
\multicolumn{10}{@{}l}{\textit{RoboMimic}} \\[2pt]
can-mh & $122{\pm}1$ & $329{\pm}6$ & $339{\pm}8$ & $389{\pm}21$ & $198{\pm}10$ & $308{\pm}7$ & $317{\pm}5$ & $289{\pm}6$ & $\mathbf{416}{\pm}16$ \\
square-mh & $40{\pm}1$ & $247{\pm}6$ & $273{\pm}13$ & $325{\pm}6$ & $113{\pm}7$ & $297{\pm}6$ & $273{\pm}3$ & $226{\pm}20$ & $\mathbf{348}{\pm}8$ \\
transport-mh & $7{\pm}1$ & $80{\pm}11$ & $86{\pm}11$ & $65{\pm}11$ & $63{\pm}5$ & $104{\pm}5$ & $90{\pm}2$ & $52{\pm}2$ & $\mathbf{105}{\pm}2$ \\
tool-hang-ph & $24{\pm}2$ & $133{\pm}3$ & $127{\pm}5$ & $112{\pm}10$ & $80{\pm}7$ & $121{\pm}12$ & $143{\pm}10$ & $128{\pm}6$ & $\mathbf{165}{\pm}3$ \\
\midrule
\multicolumn{10}{@{}l}{\textit{OGBench cube-triple}} \\[2pt]
Task 1 & $0{\pm}0$ & $1569{\pm}18$ & $1566{\pm}14$ & $838{\pm}7$ & $405{\pm}14$ & $798{\pm}18$ & $1065{\pm}371$ & $1396{\pm}10$ & $\mathbf{1582}{\pm}8$ \\
Task 2 & $0{\pm}0$ & $266{\pm}3$ & $\mathbf{329}{\pm}32$ & $91{\pm}47$ & $60{\pm}13$ & $172{\pm}3$ & $120{\pm}10$ & $134{\pm}28$ & $328{\pm}28$ \\
Task 3 & $0{\pm}0$ & $64{\pm}12$ & $109{\pm}7$ & $90{\pm}35$ & $63{\pm}5$ & $85{\pm}3$ & $117{\pm}14$ & $2{\pm}1$ & $\mathbf{262}{\pm}14$ \\
Task 4 & $0{\pm}0$ & $184{\pm}61$ & $334{\pm}16$ & $175{\pm}10$ & $33{\pm}4$ & $32{\pm}20$ & $110{\pm}1$ & $0{\pm}0$ & $\mathbf{344}{\pm}57$ \\
Task 5 & $0{\pm}0$ & $0{\pm}0$ & $66{\pm}36$ & $0{\pm}0$ & $7{\pm}7$ & $1{\pm}0$ & $3{\pm}2$ & $0{\pm}0$ & $\mathbf{174}{\pm}15$ \\
\midrule
\multicolumn{10}{@{}l}{\textit{OCBench Lite}} \\[2pt]
Block 1 & $32{\pm}6$ & $\mathbf{293}{\pm}8$ & $273{\pm}4$ & $228{\pm}11$ & $133{\pm}5$ & $137{\pm}7$ & $229{\pm}15$ & $152{\pm}13$ & $279{\pm}10$ \\
Block 2 & $37{\pm}2$ & $327{\pm}17$ & $340{\pm}42$ & $184{\pm}37$ & $185{\pm}19$ & $243{\pm}25$ & $264{\pm}3$ & $\mathbf{426}{\pm}42$ & $364{\pm}24$ \\
Chamber 1 & $80{\pm}5$ & $1302{\pm}33$ & $1269{\pm}13$ & $1817{\pm}18$ & $161{\pm}6$ & $157{\pm}9$ & $1805{\pm}18$ & $695{\pm}3$ & $\mathbf{2311}{\pm}18$ \\
Chamber 2 & $28{\pm}4$ & $\mathbf{406}{\pm}22$ & $316{\pm}47$ & $252{\pm}26$ & $78{\pm}8$ & $73{\pm}10$ & $300{\pm}2$ & $180{\pm}8$ & $388{\pm}35$ \\
Chamber 3 & $20{\pm}2$ & $464{\pm}20$ & $460{\pm}47$ & $286{\pm}22$ & $88{\pm}6$ & $80{\pm}5$ & $317{\pm}39$ & $324{\pm}11$ & $\mathbf{522}{\pm}32$ \\
\midrule
\multicolumn{10}{@{}l}{\textit{LIBERO}} \\[2pt]
Task 14 & $33$ & $17{\pm}9$ & $375{\pm}30$ & $314{\pm}42$ & $70{\pm}13$ & $\mathbf{514}{\pm}72$ & $270{\pm}42$ & $69{\pm}6$ & $470{\pm}39$ \\
Task 29 & $141$ & $122{\pm}28$ & $209{\pm}38$ & $163{\pm}19$ & $131{\pm}11$ & $\mathbf{263}{\pm}55$ & $242{\pm}25$ & $136{\pm}16$ & $168{\pm}28$ \\
Task 54 & $213$ & $119{\pm}9$ & $263{\pm}6$ & $256{\pm}5$ & $281{\pm}6$ & $\mathbf{604}{\pm}43$ & $263{\pm}11$ & $281{\pm}4$ & $572{\pm}51$ \\
Task 60 & $156$ & $548{\pm}60$ & $596{\pm}15$ & $594{\pm}44$ & $287{\pm}34$ & $537{\pm}12$ & $672{\pm}27$ & $\mathbf{709}{\pm}19$ & $630{\pm}38$ \\
Task 64 & $84$ & $156{\pm}17$ & $255{\pm}15$ & $158{\pm}11$ & $175{\pm}9$ & $118{\pm}6$ & $193{\pm}7$ & $\mathbf{274}{\pm}16$ & $249{\pm}11$ \\
\bottomrule
\end{tabular}
\caption{\textbf{Absolute throughput (successes/hour) across all tasks and methods} assuming 10\,Hz for RoboMimic, OGBench, and OCBench, and 20\,Hz for LIBERO. Values are mean $\pm$ SEM.}
\label{tab:results-tp}
\end{table*}

\FloatBarrier

\subsection{Hyperparameters}
\label{app:hyperparameters}
We include all hyper parameter details for our method and what we sweep over in table \ref{tab:gsbr-search}, the baseline sweeps in table \ref{tab:baseline-search}, and the training setup shared parameters in table \ref{tab:training-settings}.
\providecommand{\ours}{\texttt{GSBR}}

\begin{table}[!htbp]
\centering
\fontsize{9}{10}\selectfont
\setlength{\tabcolsep}{3pt}
\renewcommand{\arraystretch}{1.0}
\begin{tabular*}{\linewidth}{@{\extracolsep{\fill}}lp{.265\linewidth}lcccc@{}}
\toprule
Benchmark & Tasks & Online Env Steps & UTD & $H$ & $\gamma$ & TD clip \\
\midrule
RoboMimic & \texttt{can}, \texttt{square} & 1M & 2 & 5 & $0.998$ & $[0,1]$ \\
RoboMimic & \texttt{transport}, \texttt{tool-hang} & 2M & 2 & 5 & $0.998$ & $[0,1]$ \\
OGBench & \texttt{cube-triple-T[1-5]} & 2M & 2 & 5 & $0.99$ & None \\
OCBench Lite & \texttt{block-double-T[1-2]} \texttt{chamber-easy-T[1-3]} & 1M & 2 & 3 & $0.99$ & $[0,1]$ \\
LIBERO & \texttt{Task-\{14, 29, 54, 60, 64\}} & 1,000 episodes & 2 & 10 & $0.975$ per chunk & None \\
\bottomrule
\end{tabular*}
\caption{Training settings. $H$ is the action chunk length. UTD is per environment step; the discount $\gamma$ is per environment step except for LIBERO, where it is per chunk.}\label{tab:training-settings}
\end{table}

\begin{table}[!htbp]
\centering
\fontsize{9}{10}\selectfont
\setlength{\tabcolsep}{3pt}
\renewcommand{\arraystretch}{1.0}
\begin{tabular*}{\linewidth}{@{\extracolsep{\fill}}lll@{}}
\toprule
Task & BC weight $\alpha$ & Quantile $q$ \\
\midrule
\multicolumn{3}{l}{\textit{RoboMimic}}\\*
can-mh & $\{\mathbf{0.1},0.3,1,3,10\}$ & $\{0.1,0.3,\mathbf{0.5},0.7,0.9\}$ \\
square-mh & $\{\mathbf{3},10,30\}$ & $\{0.7,\mathbf{0.9}\}$ \\
transport-mh & $\{0.1,0.3, 1,\mathbf{3},10,30\}$ & $\{\mathbf{0.1},0.3, 0.7,0.9\}$ \\
tool-hang-ph & $\{3,\mathbf{10},30\}$ & $\{0.1,0.3,0.5,0.7,\mathbf{0.9}\}$ \\
\addlinespace[2pt]
\multicolumn{3}{l}{\textit{OGBench}}\\*
cube 1 & $\{1,\mathbf{3},10\}$ & $\{0.5,\mathbf{0.7},0.9\}$ \\
cube 2 & $\{3,10,\mathbf{30},100\}$ & $\{0.1,0.3,0.5,\mathbf{0.7},0.9\}$ \\
cube 3 & $\{30,100,300,\mathbf{1000},3000,10000\}$ & $\{0.7,\mathbf{0.9}\}$ \\
cube 4 & $\{3,10,\mathbf{30},100,300,1000\}$ & $\{0.7,\mathbf{0.9}\}$ \\
cube 5 & $\{100, 300,\mathbf{1000}\}$ & $\{0.1,0.3, 0.5,\mathbf{0.7},0.9\}$ \\
\addlinespace[2pt]
\multicolumn{3}{l}{\textit{OCBench Lite}}\\*
block 1 & $\{0.1,\mathbf{0.3},1,3,10,30\}$ & $\{0.1,0.3,0.5,0.7,\mathbf{0.9}\}$ \\
block 2 & $\{0.1,0.3,1,3,\mathbf{10}\}$ & $\{0.1,\mathbf{0.3},0.5,0.7,0.9\}$ \\
chamber 1 & $\{\mathbf{0.01},0.03,0.1\}$ & $\{\mathbf{0.1},0.3,0.5,0.7,0.9\}$ \\
chamber 2 & $\{0.01,0.03,\mathbf{0.1},0.3,1,3\}$ & $\{\mathbf{0.1},0.3,0.5,0.7,0.9\}$ \\
chamber 3 & $\{0.01,0.03,\mathbf{0.1},0.3,1\}$ & $\{0.1,0.3,\mathbf{0.5},0.7,0.9\}$ \\
\addlinespace[2pt]
\multicolumn{3}{l}{\textit{LIBERO}}\\*
LIBERO 14 & $\{30,\mathbf{100},300\}$ & $\{\mathbf{0.1},0.5,0.9\}$ \\
LIBERO 29 & $\{10,30,\mathbf{100},300,1000\}$ & $\{0.1,\mathbf{0.5},0.9\}$ \\
LIBERO 54 & $\{30,\mathbf{100},300\}$ & $\{\mathbf{0.1},0.5,0.9\}$ \\
LIBERO 60 & $\{30,\mathbf{100},300\}$ & $\{\mathbf{0.1},0.5,0.9\}$ \\
LIBERO 64 & $\{100,\mathbf{300},1000,3000\}$ & $\{\mathbf{0.1},0.5,0.9\}$ \\
\bottomrule
\end{tabular*}
\caption{\ours{} Hyperparameter sweep range.}\label{tab:gsbr-search}
\end{table}

\par

\begin{table}[!p]
\centering
\fontsize{8}{9}\selectfont
\setlength{\tabcolsep}{3pt}
\renewcommand{\arraystretch}{1.0}
\begin{tabular}{@{}l>{\raggedright\arraybackslash}p{.32\linewidth}>{\raggedright\arraybackslash}p{.52\linewidth}@{}}
\toprule
Method & Tasks & Hyperparameter sets \\
\midrule
DICE-RL & can-mh & $\alpha\in \{\mathbf{3},\allowbreak 10,\allowbreak 30,\allowbreak 100,\allowbreak 300\}$; $u\in \{\mathbf{0.3125},\allowbreak 2\}$ \\
 & square-mh & $\alpha\in \{3,\allowbreak \mathbf{10},\allowbreak 30,\allowbreak 100,\allowbreak 300\}$; $u\in \{0.3125,\allowbreak \mathbf{2}\}$ \\
 & transport-mh & $\alpha\in \{3,\allowbreak 10,\allowbreak 30,\allowbreak 100,\allowbreak \mathbf{300}\}$; $u\in \{\mathbf{0.3125},\allowbreak 2\}$ \\
 & tool-hang-ph & $\alpha\in \{\mathbf{3},\allowbreak 10,\allowbreak 30,\allowbreak 100,\allowbreak 300\}$; $u\in \{0.3125,\allowbreak \mathbf{2}\}$ \\
 & cube 1 & $\alpha\in \{1,\allowbreak 3,\allowbreak \mathbf{10},\allowbreak 30,\allowbreak 100,\allowbreak 300\}$; $u\in \{0.3125,\allowbreak \mathbf{2}\}$ \\
 & cube 2 & $\alpha\in \{1,\allowbreak 3,\allowbreak 10,\allowbreak \mathbf{30},\allowbreak 100,\allowbreak 300\}$; $u\in \{\mathbf{0.3125},\allowbreak 2\}$ \\
 & cube 3 & $\alpha\in \{1,\allowbreak 3,\allowbreak 10,\allowbreak 30,\allowbreak 100,\allowbreak \mathbf{300}\}$; $u\in \{\mathbf{0.3125},\allowbreak 2\}$ \\
 & cube 4 & $\alpha\in \{1,\allowbreak 3,\allowbreak 10,\allowbreak 30,\allowbreak 100,\allowbreak \mathbf{300}\}$; $u\in \{0.3125,\allowbreak \mathbf{2}\}$ \\
 & cube 5 & $\alpha\in \{1,\allowbreak 3,\allowbreak \mathbf{10},\allowbreak 30,\allowbreak 100,\allowbreak 300\}$; $u\in \{\mathbf{0.3125}\}$ \\
 & block 1 & $\alpha\in \{0.3,\allowbreak 1,\allowbreak 3,\allowbreak 10,\allowbreak \mathbf{30},\allowbreak 100\}$; $u\in \{\mathbf{0.3125},\allowbreak 2\}$ \\
 & block 2 & $\alpha\in \{0.3,\allowbreak \mathbf{1},\allowbreak 3,\allowbreak 10,\allowbreak 30,\allowbreak 100\}$; $u\in \{\mathbf{0.3125},\allowbreak 2\}$ \\
 & chamber 1, 3 & $\alpha\in \{\mathbf{0.3},\allowbreak 1,\allowbreak 3,\allowbreak 10,\allowbreak 30,\allowbreak 100\}$; $u\in \{0.3125,\allowbreak \mathbf{2}\}$ \\
 & chamber 2 & $\alpha\in \{0.3,\allowbreak \mathbf{1},\allowbreak 3,\allowbreak 10,\allowbreak 30,\allowbreak 100\}$; $u\in \{0.3125,\allowbreak \mathbf{2}\}$ \\
 & LIBERO 14, 29, 54 & $\alpha\in \{10,\allowbreak 30,\allowbreak 100,\allowbreak \mathbf{300}\}$; $u\in \{\mathbf{2}\}$ \\
 & LIBERO 60, 64 & $\alpha\in \{\mathbf{10},\allowbreak 30,\allowbreak 100,\allowbreak 300\}$; $u\in \{\mathbf{2}\}$ \\
\addlinespace[2pt]
EXPO-FT & can-mh; LIBERO 64 & $s\in \{0.01,\allowbreak \mathbf{0.03},\allowbreak 0.1,\allowbreak 0.3,\allowbreak 1\}$ \\
 & square-mh; cube 2 & $s\in \{0.01,\allowbreak \mathbf{0.03},\allowbreak 0.1,\allowbreak 0.3\}$ \\
 & transport-mh & $s\in \{\mathbf{0.01},\allowbreak 0.03,\allowbreak 0.1,\allowbreak 0.3\}$ \\
 & tool-hang-ph; cube 3--5 & $s\in \{0.01,\allowbreak 0.03,\allowbreak \mathbf{0.1},\allowbreak 0.3\}$ \\
 & cube 1; chamber 1--2 & $s\in \{0.01,\allowbreak 0.03,\allowbreak 0.1,\allowbreak 0.3,\allowbreak \mathbf{1}\}$ \\
 & block 1--2; chamber 3; LIBERO 29 & $s\in \{0.01,\allowbreak 0.03,\allowbreak 0.1,\allowbreak \mathbf{0.3},\allowbreak 1\}$ \\
 & LIBERO 14, 60 & $s\in \{0.01,\allowbreak 0.03,\allowbreak \mathbf{0.1},\allowbreak 0.3,\allowbreak 1\}$ \\
 & LIBERO 54 & $s\in \{\mathbf{0.01},\allowbreak 0.03,\allowbreak 0.1,\allowbreak 0.3,\allowbreak 1\}$ \\
\addlinespace[2pt]
Residual & can-mh; cube 1--4; chamber 2--3 & $s\in \{0.01,\allowbreak 0.03,\allowbreak 0.1,\allowbreak \mathbf{0.3},\allowbreak 1\}$ \\
 & square-mh & $s\in \{0.01,\allowbreak 0.03,\allowbreak \mathbf{0.1},\allowbreak 0.3\}$ \\
 & transport-mh; tool-hang-ph & $s\in \{0.01,\allowbreak \mathbf{0.03},\allowbreak 0.1,\allowbreak 0.3\}$ \\
 & cube 5 & $s\in \{0.01,\allowbreak 0.03,\allowbreak 0.1,\allowbreak 0.3\}$ (all tied) \\
 & block 1; chamber 1 & $s\in \{0.01,\allowbreak 0.03,\allowbreak 0.1,\allowbreak 0.3,\allowbreak \mathbf{1}\}$ \\
 & block 2; LIBERO 64 & $s\in \{0.01,\allowbreak \mathbf{0.03},\allowbreak 0.1,\allowbreak 0.3,\allowbreak 1\}$ \\
 & LIBERO 14, 29, 60 & $s\in \{0.01,\allowbreak 0.03,\allowbreak \mathbf{0.1},\allowbreak 0.3,\allowbreak 1\}$ \\
 & LIBERO 54 & $s\in \{\mathbf{0.01},\allowbreak 0.03,\allowbreak 0.1,\allowbreak 0.3,\allowbreak 1\}$ \\
\addlinespace[2pt]
DSRL & can-mh; square-mh; block 2; chamber 1--3 & $b\in \{1,\allowbreak 1.5,\allowbreak \mathbf{2}\}$ \\
 & transport-mh; cube 3, 5; LIBERO 60, 64 & $b\in \{\mathbf{1},\allowbreak 1.5,\allowbreak 2\}$ \\
 & tool-hang-ph; cube 1--2, 4; block 1; LIBERO 14, 29, 54 & $b\in \{1,\allowbreak \mathbf{1.5},\allowbreak 2\}$ \\
\addlinespace[2pt]
RLT & can-mh & $\alpha\in \{0.03,\allowbreak 0.1,\allowbreak 0.3,\allowbreak \mathbf{1},\allowbreak 3\}$ \\
 & square-mh; transport-mh; tool-hang-ph & $\alpha\in \{0.1,\allowbreak 0.3,\allowbreak \mathbf{1},\allowbreak 3,\allowbreak 10,\allowbreak 30,\allowbreak 100,\allowbreak 300,\allowbreak 1000\}$ \\
 & cube 1 & $\alpha\in \{1,\allowbreak \mathbf{3},\allowbreak 10,\allowbreak 30,\allowbreak 100,\allowbreak 300,\allowbreak 1000\}$ \\
 & cube 2, 4 & $\alpha\in \{1,\allowbreak 3,\allowbreak 10,\allowbreak \mathbf{30},\allowbreak 100,\allowbreak 300,\allowbreak 1000\}$ \\
 & cube 3, 5 & $\alpha\in \{1,\allowbreak 3,\allowbreak 10,\allowbreak 30,\allowbreak \mathbf{100},\allowbreak 300,\allowbreak 1000\}$ \\
 & block 1; chamber 1--3 & $\alpha\in \{\mathbf{0.1},\allowbreak 0.3,\allowbreak 1,\allowbreak 3,\allowbreak 10,\allowbreak 30,\allowbreak 100,\allowbreak 300,\allowbreak 1000\}$ \\
 & block 2 & $\alpha\in \{1,\allowbreak 3,\allowbreak \mathbf{10},\allowbreak 30,\allowbreak 100,\allowbreak 300,\allowbreak 1000\}$ \\
 & LIBERO 14, 29, 64 & $\alpha\in \{10, 30, \mathbf{100}, 300, 1000\}$ \\
 & LIBERO 54 & $\alpha\in \{10, 30, 100, \mathbf{300},1000\}$ \\
 & LIBERO 60 & $\alpha\in \{10, \mathbf{30},100, 300, 1000\}$  \\
\addlinespace[2pt]
TD3+BC & can-mh & $\alpha\in \{0.03,\allowbreak 0.1,\allowbreak 0.3,\allowbreak \mathbf{1},\allowbreak 3\}$ \\
 & square-mh; cube 1 & $\alpha\in \{1,\allowbreak \mathbf{3},\allowbreak 10,\allowbreak 30,\allowbreak 100,\allowbreak 300,\allowbreak 1000\}$ \\
 & transport-mh & $\alpha\in \{0.1,\allowbreak 0.3,\allowbreak 1,\allowbreak \mathbf{3},\allowbreak 10,\allowbreak 30,\allowbreak 100,\allowbreak 300,\allowbreak 1000\}$ \\
 & tool-hang-ph & $\alpha\in \{0.1,\allowbreak 0.3,\allowbreak \mathbf{1},\allowbreak 3,\allowbreak 10,\allowbreak 30,\allowbreak 100,\allowbreak 300,\allowbreak 1000\}$ \\
 & cube 2, 4 & $\alpha\in \{0.1,\allowbreak 0.3,\allowbreak 1,\allowbreak 3,\allowbreak 10,\allowbreak \mathbf{30},\allowbreak 100,\allowbreak 300,\allowbreak 1000\}$ \\
 & cube 3 & $\alpha\in \{0.1,\allowbreak 0.3,\allowbreak 1,\allowbreak 3,\allowbreak 10,\allowbreak 30,\allowbreak \mathbf{100},\allowbreak 300,\allowbreak 1000\}$ \\
 & cube 5 & $\alpha\in \{0.1,\allowbreak 0.3,\allowbreak 1,\allowbreak 3,\allowbreak 10,\allowbreak 30,\allowbreak 100,\allowbreak 300,\allowbreak 1000\}$ (all tied) \\
 & block 1; chamber 1--3 & $\alpha\in \{\mathbf{0.1},\allowbreak 0.3,\allowbreak 1,\allowbreak 3,\allowbreak 10,\allowbreak 30,\allowbreak 100,\allowbreak 300,\allowbreak 1000\}$ \\
 & block 2 & $\alpha\in \{1,\allowbreak 3,\allowbreak \mathbf{10},\allowbreak 30,\allowbreak 100,\allowbreak 300,\allowbreak 1000\}$ \\
 & LIBERO 14, 29, 64 & $\alpha\in \{10, 30, 100, \mathbf{300}, 1000\}$ \\
 & LIBERO 54 & $\alpha\in \{10, 30, 100, 300, \mathbf{1000}\}$ \\
 & LIBERO 60 & $\alpha\in \{10, 30, \mathbf{100}, 300, 1000\}$ \\
\bottomrule
\end{tabular}
\caption{Baseline search grids. Bold values are selected. Here $s$ is residual scale, $b$ is the DSRL bound/noise scale, and $u$ is UTD.}\label{tab:baseline-search}\label{tab:baseline-selected}

\vspace{3pt}
\begin{minipage}{\linewidth}
\end{minipage}
\end{table}

\FloatBarrier

\clearpage
\subsection{Sensitivity Analysis}
\begin{figure}[!ht]
    \centering
    \includegraphics[width=\linewidth,height=0.74\textheight,keepaspectratio]{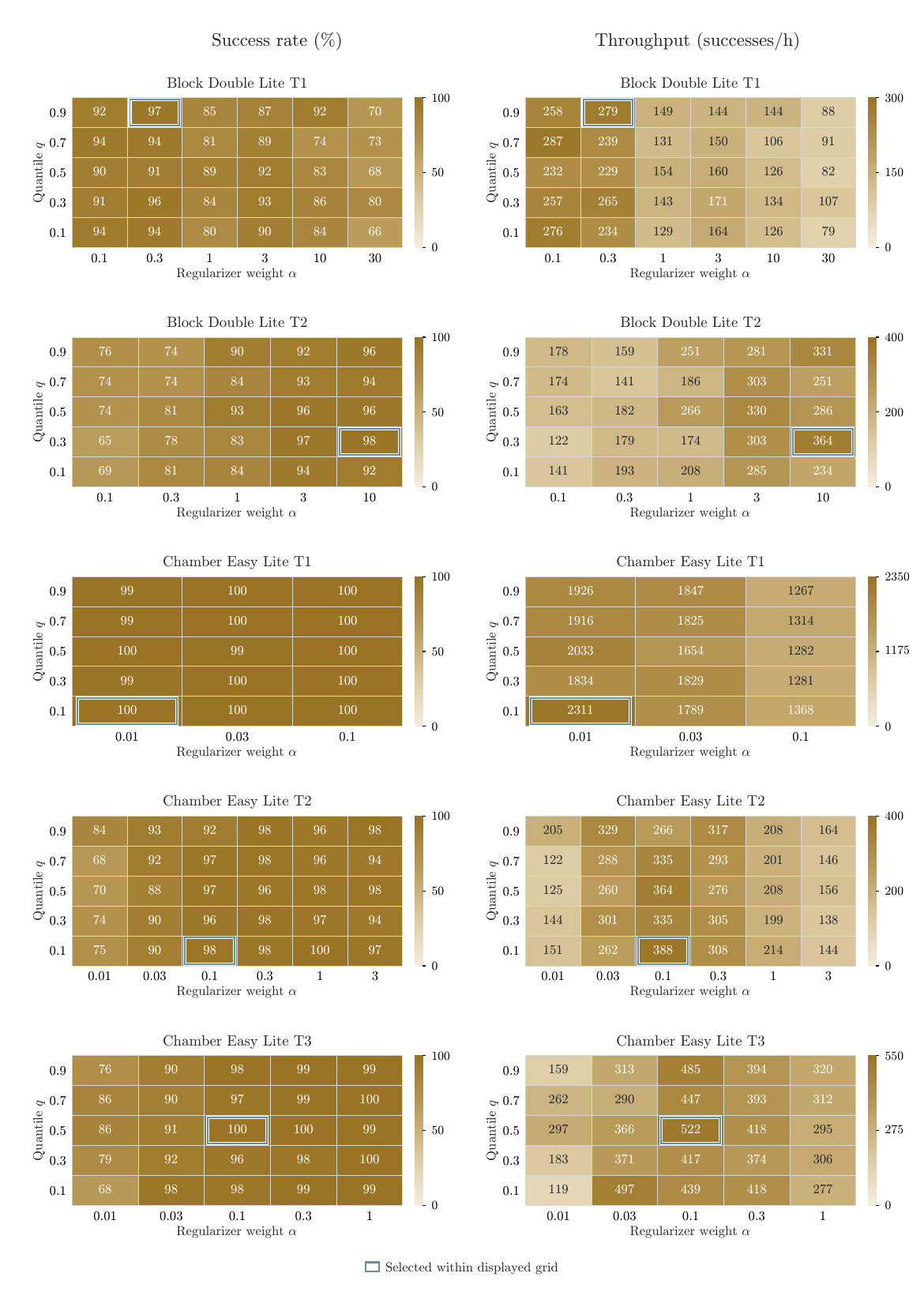}
    \caption{This figure demonstrates how the success rate, and throughput vary across different hyper parameter configs. Our method performs well across multiple magnitudes of BC weights.}
    \label{fig:sensitivity}
\end{figure}

We visualize the hyper parameter grid for our method in figure \ref{fig:sensitivity}. Our method across most tasks under a specific quantile performs similarly across different magnitudes of behavior constraint weights.

\end{document}